\documentclass{article}

\usepackage[
   top=25mm,
   bottom=25mm,
   left=30mm,
   right=30mm]{geometry}

\usepackage{amsmath,amsfonts,amssymb}
\usepackage{bm}
\usepackage{array}
\usepackage[caption=false,font=normalsize,labelfont=sf,textfont=sf]{subfig}
\usepackage{textcomp}
\usepackage{stfloats}
\usepackage{url}
\usepackage{verbatim}
\usepackage{graphicx}
\usepackage{cite}
\usepackage{tabularx}
\usepackage{booktabs}
\usepackage[ruled,vlined,linesnumbered]{algorithm2e}

\begin{document}

\title{On the Derivation of Tightly-Coupled LiDAR-Inertial Odometry with VoxelMap}
\author{Zhihao Zhan\thanks{Email: zhihazhan2-c@my.cityu.edu.hk}}
\date{}
\maketitle

\begin{abstract}
This note presents a concise mathematical formulation of tightly-coupled LiDAR-Inertial Odometry within an iterated error-state Kalman filter framework using a VoxelMap representation. 
Rather than proposing a new algorithm, it provides a clear and self-contained derivation that unifies the geometric modeling and probabilistic state estimation through consistent notation and explicit formulations.
The document is intended to serve both as a technical reference and as an accessible entry point for a foundational understanding of the system architecture and estimation principles.
\end{abstract}

\section{Introduction}
\label{sec:intro}
The rapid development of 3D Light Detection and Ranging (LiDAR) technology has accelerated its adoption across diverse applications, including autonomous driving, robotics, and large-scale environmental mapping~\cite{nguyen2022ijrr-ntuviral,li2024ijrr-marslvig,zhan2025agrilira4d}. 
Owing to its high geometric accuracy and robustness to illumination variations, LiDAR has become a primary sensing modality for perception and state estimation, driving significant progress in LiDAR-based odometry, particularly LiDAR–Inertial Odometry (LIO), which integrates inertial measurements for accurate and robust motion estimation.

To improve estimation robustness and consistency, modern LIO systems predominantly adopt tightly-coupled formulations, where inertial and LiDAR measurements are jointly incorporated within a unified estimation framework~\cite{fastlio2,fasterlio,liosam2020shan,liliom}. 
By directly integrating geometric constraints into the filtering or optimization process, tightly-coupled approaches enable consistent uncertainty propagation and stronger cross-modal coupling than loosely-coupled strategies.
Error-state Kalman filter (ESKF) and its iterative variants are therefore widely adopted due to their computational efficiency and suitability for real-time estimation~\cite{sola2017quaternionkinematicserrorstatekalman}.
In particular, the iterated extended Kalman filter (IEKF) further enhances the ESKF framework by repeatedly relinearizing nonlinear measurement models around the latest estimate, making it especially suitable for handling the strong nonlinearity of LiDAR geometric constraints~\cite{ikf_as_gn}.

In tightly-coupled LIO systems, the effectiveness of geometric constraints depends strongly on how the environment is represented during estimation. 
The map representation must preserve local structure while supporting efficient data association and incremental updates under real-time constraints. 
Traditional feature-based or dense representations often struggle to balance accuracy and efficiency, motivating structured spatial representations. 
Among these, voxel-based maps have gained increasing attention for organizing measurements into locally consistent regions while maintaining computational scalability, making them well suited for tightly-coupled LIO systems~\cite{yuan2022efficient,wu2023voxelmap++,yang2024c,liu2026voxel}.

Despite the rapid development of tightly-coupled LIO systems, their underlying formulations are often presented in implementation-oriented forms, where geometric modeling and estimation procedures are described separately or implicitly. 
As a result, the interaction between probabilistic map representation and state estimation is not always explicitly articulated. 
Motivated by this observation, this note provides a concise and self-contained mathematical derivation of VoxelMap-based tightly-coupled LIO with consistent notation and explicit formulation.

The remainder of this note is organized as follows. 
Section~\ref{sec:preliminaries} introduces the system notation, operators, and kinematic modeling. 
Section~\ref{sec:voxelmap} presents the probabilistic VoxelMap~\cite{yuan2022efficient} representation and its construction. 
Section~\ref{sec:lio} derives the tightly-coupled LIO formulation, including state propagation, motion compensation, geometric residual modeling, and iterated Kalman updates, following a pipeline consistent with FAST-LIO2~\cite{fastlio2}. 
Section~\ref{sec:conclusion} concludes the note.

\section{Preliminaries}
\label{sec:preliminaries}
\begin{figure*}[htbp]
    \centering
    \includegraphics[width=0.8\textwidth]{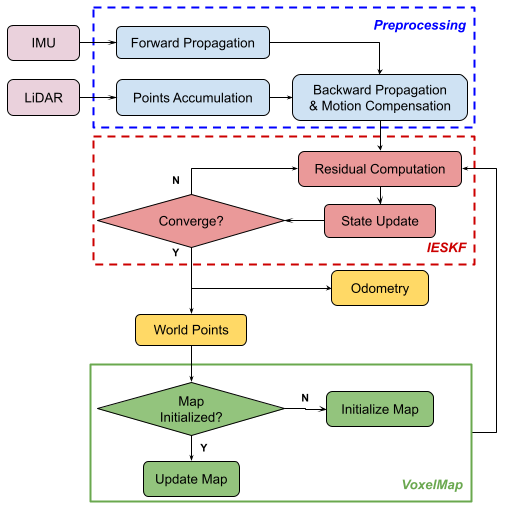}
    \caption{\textbf{System overview of the VoxelMap-based LIO framework.} 
    LiDAR and IMU measurements are preprocessed and fused within an iterated error-state Kalman filter. The estimated pose registers points to the global frame and incrementally updates the VoxelMap.}
    \label{fig:system_pipeline}
\end{figure*}

This note adopts the notation summarized in Table~\ref{tab:c_runtime}. 
An overview of the system workflow is illustrated in Figure~\ref{fig:system_pipeline}.
LiDAR and inertial measurement unit (IMU) measurements are first processed to obtain undistorted point clouds and propagated states.
The preprocessed measurements are then incorporated into an iterated error-state Kalman filter (IESKF) for state correction at the LiDAR rate (e.g., 10 Hz).
The estimated pose is used to register the LiDAR points to the global frame and update the VoxelMap constructed so far. 
The updated map subsequently serves as the reference for registering incoming measurements.

\begin{table}[htbp]
\centering
\caption{\textbf{Important Notations.}}
\label{tab:c_runtime}
    \begin{tabularx}{0.8\textwidth}{lX}
        \toprule
        Symbols & Meaning \\ 
        \midrule
        $t_k$ & The scan-end time of the $k$-th LiDAR scan.\\
        $\tau_i$ & The $i$-th IMU sample time in a LiDAR scan.\\
        $\rho_j$ & The $j$-th feature point's sample time in a LiDAR scan.\\
        $I_i, I_j, I_k$ & The IMU body frame at the time $\tau_i$, $\rho_j$ and $t_k$.\\
        $L_j, L_k$ & The LiDAR body frame at the time $\rho_j$ and $t_k$.\\
        $\mathbf{x}, \widehat{\mathbf{x}}, \bar{\mathbf{x}}$ & The ground-true, propagated, and updated value of $\mathbf{x}$. \\
        $\widetilde{\mathbf{x}}$ & The error between ground-true $\mathbf{x}$ and its estimation $\bar{\mathbf{x}}$.\\
        $\widehat{\mathbf{x}}^\kappa$ & The $\kappa$-th update of $\mathbf{x}$ in the iterated Kalman filter.\\
        ${\mathbf{x}}_{i}, \mathbf{x}_j, \mathbf{x}_k$ & The vector (e.g.,state) $\mathbf{x}$ at time $\tau_i, \rho_j$ and $t_k$.\\
        $\check{\mathbf{x}}_j$ & Estimate of $\mathbf{x}_j$ relative to $\mathbf{x}_k$ in the back propagation. \\
        \bottomrule
    \end{tabularx}
\end{table}

\subsection{Operator $\boxplus$ / $\boxminus$}
\label{sec:preli_operator}
Following~\cite{hertzberg2013integrating,fastlio2}, let $\mathcal{M}$ denote an $n$-dimensional manifold (e.g., $\mathcal{M} = \rm{SO}\left(3\right)$). 
Owing to the local homeomorphism between manifolds and $\mathbb{R}_n$, a local bijection between $\mathcal{M}$ and its tangent space can be defined using the encapsulation operators $\boxplus$ and $\boxminus$
\begin{equation}
\label{e:operator}
    \begin{aligned}
        &\ \ \ \ \ \ \ \ \ \ \ \boxplus: \mathcal{M} \times \mathbb{R}_n \rightarrow \mathcal{M}, 
        \ \ \ \ \ \ \boxminus: \mathcal{M} \times \mathcal{M} \rightarrow \mathbb{R}_n,\\
        &\mathcal{M} = \rm{SO}\left(3\right): \mathbf{R} \boxplus \mathbf{r} = \mathbf{R} \rm{Exp}\left(\mathbf{r}\right),\ \ \ \ 
        \mathbf{R}_1 \boxminus \mathbf{R}_2 = \rm{Log}\left(\mathbf{R}_2^{T}\mathbf{R}_1\right), \\
        &\mathcal{M} = \mathbb{R}_n: \ \ \ \ \ \mathbf{a} \boxplus \mathbf{b} = \mathbf{a} + \mathbf b,
        \ \ \ \ \ \ \ \ \  \mathbf{a} \boxminus \mathbf{b} = \mathbf{a} - \mathbf{b}.
    \end{aligned}
\end{equation}
where 
$\textstyle\text{Exp}\left(\mathbf{r}\right) = \mathbf{I} + \frac{\lfloor\mathbf{r}\rfloor_{\wedge}}{\|\mathbf{r}\|} \sin\left(\|\mathbf{r}\|\right) + \frac{\lfloor\mathbf{r}\rfloor_{\wedge}^2}{\|\mathbf{r}\|^2} \left(1 - \cos\left(\|\mathbf{r}\|\right) \right)$ is the exponential map and $\text{Log}(\cdot)$ is its inverse map.
For a compound manifold $\mathcal{M} = \rm{SO}\left(3\right) \times \mathbb{R}_n$ we have
$$
\begin{bmatrix} \mathbf{R} \\ \mathbf{a} \end{bmatrix} \boxplus \begin{bmatrix} \mathbf{r} \\ \mathbf{b} \end{bmatrix} = \begin{bmatrix} \mathbf{R} \boxplus \mathbf{r} \\ \mathbf{a} + \mathbf{b} \end{bmatrix}, \ \ 
\begin{bmatrix} \mathbf{R}_1 \\ \mathbf{a} \end{bmatrix} \boxminus \begin{bmatrix}\mathbf{R}_2 \\ \mathbf{b} \end{bmatrix} = \begin{bmatrix} \mathbf{R}_1 \boxminus \mathbf{R}_2 \\ \mathbf{a} - \mathbf{b} \end{bmatrix}.
$$
From the above definition, it is easy to verify that
$$
\left(\mathbf{x} \boxplus \mathbf{u} \right)  \boxminus \mathbf{x} = \mathbf{u}, \ 
\mathbf{x} \boxplus \left(\mathbf{y} \boxminus \mathbf{x}\right) = \mathbf{y}, \ \forall \mathbf{x}, \mathbf{y} \in \mathcal{M}, \ \forall \mathbf{u} \in \mathbb{R}_n. 
$$

\subsection{Kinematic Models}
\label{sec:preli_kinematics}
The evolution of the system state is governed by a kinematic model driven by inertial measurements. 
For clarity of derivation, the dynamics are first formulated in continuous time to capture the underlying motion. 
The model is then discretized over finite sampling intervals, yielding a discrete-time representation used in the subsequent estimation framework.

\subsubsection{Continuous Model}
Assuming that the IMU is rigidly attached to the LiDAR with a known extrinsic ${}^I_L \mathbf{T} = \left({}^I_L \mathbf{R}, {}^I_L \mathbf{t}\right)$, we take the IMU frame (denoted as $\{I\}$) as the body frame of reference. 
The global frame $\{G\}$ is chosen to coincide with the initial IMU frame. 
This leads to the following continuous-time kinematic model
\begin{equation}
\label{e:kine_model_continuous}
    \begin{aligned}
        {}^G_I \dot{\mathbf{t}} 
        &= {}^G_I \mathbf{v},
        \ {}^G_I \dot{\mathbf{v}} = {}^G_I \mathbf{R} \left( \mathbf{a}_m - \mathbf{b}_{\mathbf{a}} - \mathbf{n}_{\mathbf{a}} \right) + {}^G \mathbf{g},
        \ {}^G \dot{\mathbf{g}} = \mathbf{0}, \\
        {}^G_I \dot{\mathbf{R}} 
        &= {}^G_I \mathbf{R} \lfloor \bm{\omega}_m -  \mathbf{b}_{\bm{\omega}} - \mathbf{n}_{\bm{\omega}} \rfloor_{\wedge},
        \ \dot{\mathbf{b}}_{\bm{\omega}} = \mathbf{n}_{\mathbf{b}\bm{\omega}}, 
        \ \dot{\mathbf{b}}_{\mathbf{a}} = \mathbf{n}_{\mathbf{b} \mathbf{a}}
    \end{aligned}
\end{equation}
where ${}^G_I \mathbf{t}$, ${}^G_I \mathbf{R}$ are the position and attitude of IMU in the global frame $\{G\}$, $ {}^G \mathbf{g}$ is the gravity vector expressed in $\{G\}$, $\mathbf{a}_m$ and $\bm{\omega}_m$ are IMU measurements, $\mathbf{n}_{\mathbf{a}}$ and $\mathbf{n}_{\bm{\omega}}$ are zero-mean white Gaussian measurement noises, $\mathbf{b}_{\mathbf{a}}$ and $\mathbf{b}_{\bm{\omega}}$ are IMU biases modeled as random walks driven by Gaussian noises $\mathbf{n}_{\mathbf{b} \mathbf{a}}$ and $\mathbf{n}_{\mathbf{b} \bm{\omega}}$, and the notation $\lfloor \mathbf{a} \rfloor_{\wedge}$ denotes the skew-symmetric matrix of vector $\mathbf{a} \in \mathbb{R}_3$ that maps the cross product operation.

The continuous-time system dynamics in (\ref{e:kine_model_continuous}) are derived from the following IMU measurement model
\begin{equation}
\label{e:imu_model}
    \begin{aligned}
        \mathbf{a}_m 
        &= {}^G_I \mathbf{R}^T \left({}^G \mathbf{a} - {}^G \mathbf{g}\right) + \mathbf{b}_{\mathbf{a}} + \mathbf{n}_{\mathbf{a}}, \\
        \bm{\omega}_m 
        &= \bm{\omega} + \mathbf{b}_{\bm{\omega}} + \mathbf{n}_{\bm{\omega}}, \\
        \dot{\mathbf{b}}_{\bm{\omega}} 
        &= \mathbf{n}_{\mathbf{b}\bm{\omega}}, 
        \ \dot{\mathbf{b}}_{\mathbf{a}} = \mathbf{n}_{\mathbf{b} \mathbf{a}}
    \end{aligned}
\end{equation}
where ${}^G\mathbf{a}$ denotes the true linear acceleration of the IMU expressed in the global frame, and $\bm{\omega}$ is the true angular velocity expressed in the IMU frame.

Based on the IMU measurement model in (\ref{e:imu_model}), the position dynamics follow directly from the definition of velocity, i.e., ${}^G_I \dot{\mathbf{t}} = {}^G_I \mathbf{v}$. 

The velocity dynamics are obtained by rewriting the accelerometer measurement equation as below.
\begin{equation}
\label{e:vel_dyn}
    \begin{aligned}
        \mathbf{a}_m - \mathbf{b}_{\mathbf{a}} - \mathbf{n}_{\mathbf{a}} 
        &= {}^G_I \mathbf{R}^T \left({}^G \mathbf{a} - {}^G \mathbf{g}\right), \\
        \to {}^G \mathbf{a} 
        &= {}^G_I \mathbf{R}\left(\mathbf{a}_m - \mathbf{b}_{\mathbf{a}} - \mathbf{n}_{\mathbf{a}}\right) + {}^G \mathbf{g}.
    \end{aligned}
\end{equation}
Since the true linear acceleration in the global frame satisfies ${}^G \mathbf{a} = {}^G_I\dot{\mathbf{v}}$, substituting into the above expression yields ${}^G_I \dot{\mathbf{v}} = {}^G_I \mathbf{R}\left(\mathbf{a}_m - \mathbf{b}_{\mathbf{a}} - \mathbf{n}_{\mathbf{a}}\right) + {}^G \mathbf{g}$.

Similarly, the attitude kinematics on $SO\left(3\right)$ are governed by ${}^G_I \dot{\mathbf{R}} = {}^G_I \mathbf{R}\lfloor \bm{\omega} \rfloor_{\wedge}$, where $\bm{\omega}$ is the true angular velocity expressed in the IMU frame. 
Substituting the gyroscope model in (\ref{e:imu_model}), gives the continuous-time attitude dynamics
\begin{equation}
\label{e:att_dyn}
    {}^G_I \dot{\mathbf{R}} = {}^G_I \mathbf{R} \lfloor \bm{\omega}_m - \mathbf{b}_{\bm{\omega}} - \mathbf{n}_{\bm{\omega}} \rfloor_{\wedge}.
\end{equation}

The gravity vector is assumed constant in the global frame, i.e., ${}^G \dot{\mathbf{g}} = \mathbf{0}$.
Moreover, the gyroscope and accelerometer biases are modeled as random walks driven by white Gaussian noises, yielding $\dot{\mathbf{b}}_{\bm{\omega}} = \mathbf{n}_{\mathbf{b}\bm{\omega}}, \dot{\mathbf{b}}_{\mathbf{a}} = \mathbf{n}_{\mathbf{b} \mathbf{a}}$.

\subsubsection{Discrete Model}
By integrating the continuous-time kinematics over a sampling interval $\Delta t$, the following exact integral form is obtained
\begin{equation}
\label{e:integ_con}
    \begin{aligned}
        {}^G_I \mathbf{R} \left(t + \Delta t\right) 
        &= {}^G_I \mathbf{R} \left(t\right) \rm{Exp} \left( \int_t^{t + \Delta t} \bm{\omega} \left(\tau\right) d\tau \right),	\\
        {}^G_I \mathbf{v} \left(t + \Delta t\right) 
        &= {}^G_I \mathbf{v} \left(t\right) + \int_t^{t + \Delta t} {}^G \mathbf{a} \left(\tau\right) d\tau, \\
        {}^G_I \mathbf{t} \left(t + \Delta t\right) 
        &= {}^G_I \mathbf{t} \left(t\right) + \int_t^{t + \Delta t} {}^G_I \mathbf{v} \left(\tau\right) d\tau .
    \end{aligned}
\end{equation}

Starting from (\ref{e:integ_con}), a discrete-time model is obtained by assuming that the angular velocity and acceleration are approximately constant within each sampling interval $\left[t_i, t_{i+1}\right]$ with $t_{i+1} = t_i + \Delta t$. Using a first-order approximation for the integrals yields
\begin{equation}
\label{e:integ_discrete}
    \begin{aligned}
        {}^G_{I_{i+1}} \mathbf{R}
        &= {}^G_{I_i} \mathbf{R} \rm{Exp} \left(\bm{\omega}_i \Delta t\right), \\
        {}^G_{I_{i+1}} \mathbf{v}
        &= {}^G_{I_i} \mathbf{v} + {}^G\mathbf{a}_i \Delta t, \\
        {}^G_{I_{i+1}} \mathbf{t}
        &= {}^G_{I_i} \mathbf{t} + {}^G_{I_i} \mathbf{v} \Delta t
    \end{aligned}
\end{equation}
where ${}^G_{I_i} \mathbf{R} \triangleq {}^G_I \mathbf{R} \left(t_i\right)$, ${}^G_{I_i} \mathbf{v} \triangleq {}^G_I \mathbf{v}\left(t_i\right)$, and $ {}^G_{I_i} \mathbf{t} \triangleq {}^G_I \mathbf{t}\left(t_i\right)$.
Substituting the IMU measurement model in (\ref{e:imu_model}), gives the following discrete model
\begin{equation}
\label{e:integ_discrete_all}
    \begin{aligned}
        {}^G_{I_{i+1}} \mathbf{R}
        &= {}^G_{I_i} \mathbf{R} \rm{Exp} \left( \left( \bm{\omega}_{m_i} -  \mathbf{b}_{\bm{\omega}_i} - \mathbf{n}_{\bm{\omega}_i}\right) \Delta t\right), \\
        {}^G_{I_{i+1}} \mathbf{v}
        &={}^G_{I_i} \mathbf{v} + \left({}^G_{I_i} \mathbf{R} \left(\mathbf{a}_{m_i} - \mathbf{b}_{\mathbf{a}_i} - \mathbf{n}_{\mathbf{a}_i}\right) + {}^G \mathbf{g}_i \right) \Delta t, \\
        {}^G_{I_{i+1}} \mathbf{t}
        &= {}^G_{I_i} \mathbf{t} + {}^G_{I_i} \mathbf{v} \Delta t.
    \end{aligned}
\end{equation}

Based on the $\boxplus$ operation and discrete-time model in (\ref{e:integ_discrete_all}) defined above, we can discretize the continuous model in (\ref{e:kine_model_continuous}) at the IMU sampling period $\Delta t$. The resultant discrete model is
\begin{equation}
\label{e:kine_model_discrete}
    \begin{aligned}
        \mathbf{x}_{i+1} = \mathbf{x}_{i} \boxplus \left( \Delta t \mathbf{f}(\mathbf{x}_i, \mathbf{u}_i, \mathbf{w}_i) \right)
    \end{aligned}
\end{equation}
where $i$ is the index of IMU measurements, the function $\mathbf{f}$, state $\mathbf{x}$, input $\mathbf{u}$ and noise $\mathbf{w}$ are defined as below
\begin{equation}
\label{e:states_k}
    \begin{split}
        \mathcal{M} &= SO\left(3\right) \times \mathbb{R}_{15}, \ \text{dim}(\mathcal{M}) = 18, \\
        \mathbf{x} &\doteq \begin{bmatrix} {}^G_I \mathbf{R}^T & {}^G_I \mathbf{t}^T & {}^G_I \mathbf{v}^T & \mathbf{b}_{\bm{\omega}}^T & \mathbf{b}_{\mathbf{a}}^T & {}^G \mathbf{g}^T \end{bmatrix}^T \in \mathcal{M}, \\
        \mathbf{u} &\doteq \begin{bmatrix} \bm{\omega}^T_m & \mathbf{a}^T_m\end{bmatrix}^T,\
        \mathbf{w} \doteq \begin{bmatrix}\mathbf n_{\bm{\omega}}^T & \mathbf{n}_{\mathbf{a}}^T & \mathbf{n}_{\mathbf{b} \bm{\omega}}^T & \mathbf{n}_{\mathbf{b} \mathbf{a}}^T \end{bmatrix}^T, \\
        \mathbf{f} &\left( \mathbf{x}_{i}, \mathbf{u}_i, \mathbf{w}_i \right) = \begin{bmatrix} \bm{\omega}_{m_i} - \mathbf{b}_{\bm{\omega}_i} - \mathbf{n}_{\bm{\omega}_i}, \\
        {}^G_{I_i} \mathbf{v} \\
        {}^G_{I_i} \mathbf{R} \left(\mathbf{a}_{m_i} - \mathbf{b}_{\mathbf{a}_i} - \mathbf{n}_{\mathbf{a}_i}\right) + {}^G \mathbf{g}_i  \\
        \mathbf{n}_{\mathbf{b} \bm{\omega}_i} \\
        \mathbf{n}_{\mathbf{b} \mathbf{a}_i} \\
        \mathbf{0}_{3\times1} \end{bmatrix}.
    \end{split}
\end{equation}

\section{Probabilistic VoxelMap Representation}
\label{sec:voxelmap}
\subsection{Probabilistic Plane Representation}
\label{sec:voxel_prob}
Following the probabilistic VoxelMap formulation in~\cite{yuan2022efficient}, each voxel maintains a single probabilistic plane feature to represent local geometry. 
Since a plane is estimated from its associated LiDAR points, its uncertainty is determined by the uncertainties of those points. 
As VoxelMap is expressed in the global frame, the uncertainty of point ${}^G \mathbf{p}_i$ must be analyzed after transformation from the local LiDAR frame, leading to two principal noise sources: raw measurement noise and uncertainty of pose estimation. 
The following subsections derive the corresponding uncertainty propagation from points to planes and establish the probabilistic plane representation.

\subsubsection{Uncertainty of point ${}^G \mathbf{p}_i$}
\begin{figure*}[htbp]
    \centering
    \includegraphics[width=0.8\textwidth]{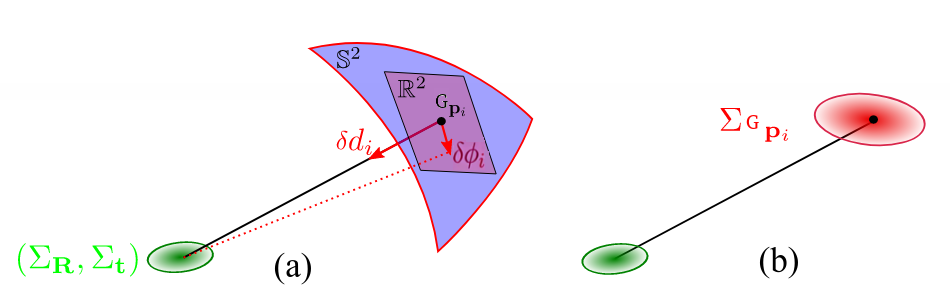}
    \caption{\textbf{The uncertainty model of the registered point.}}
    \label{fig:uncertainty_pt}
\end{figure*}

Following the LiDAR measurement noise analysis in~\cite{yuan2021pixel}, the uncertainty of a point in the local LiDAR frame is modeled as the combination of ranging noise and bearing direction noise as shown in Figure~\ref{fig:uncertainty_pt}. 
Let $\bm{\phi} \in \mathbb{S}^2$ denote the measured bearing direction with perturbation $\delta_{\bm{\phi}_i} \sim \mathcal{N}\left(\mathbf{0}_{2\times1}, \bm{\Sigma}_{\phi_i}\right)$ defined on its tangent plane, and let $d_i$ be the measured depth with ranging noise $\delta_{d_i} \sim \mathcal{N}\left(0, \Sigma_{d_i}\right)$.

Using the $\boxplus$ operation encapsulated in $\mathbb{S}^2$~\cite{he2021kalman}, define the relation between the true bearing direction $\bm{\phi}_i^{gt}$ and its measurement $\bm{\phi}_i$ as below
\begin{equation}
\label{e:uncertainty_pt_direction}
    \begin{aligned}
        \bm{\phi}_i^{gt} 
        &= \bm{\phi}_i \boxplus_{\mathbb{S}^2} \delta_{\bm{\phi}_i} \triangleq \rm{Exp} \left( \mathbf{N} \left(\bm{\phi}_i \right) \delta_{\bm{\phi}_i} \right) \bm{\phi}_i \\
        &\approx \left(I + \lfloor \mathbf{N} \left(\bm{\phi}_i \right) \delta_{\bm{\phi}_i} \rfloor_{\wedge} \right) \bm{\phi}_i
    \end{aligned}
\end{equation}
where $\mathbf{N} \left( \bm{\phi}_i \right) = \begin{bmatrix} \mathbf{N}_1 & \mathbf{N}_2 \end{bmatrix} \in \mathbb{R}_{3\times2}$ is an orthonormal basis of the tangent plane at $\bm{\phi}_i$ (see Figure~\ref{fig:uncertainty_pt}).
The $\boxplus_{\mathbb{S}^2}$ operation essentially rotates the unit vector $\bm{\phi}_i$ about the axis $\delta_{\bm{\phi}_i}$ in the tangent plane at $\bm{\phi}_i$, the result is still a unit vector (i.e., remain on $\mathbb{S}^2$).

Similarly, the ground-truth depth $d^{gt}_i$ can be expressed below
\begin{equation}
\label{e:uncertainty_pt_depth}
    \begin{aligned}
        d^{gt}_i = d_i + \delta_{d_i}.
    \end{aligned}
\end{equation}

Combining (\ref{e:uncertainty_pt_direction}) and (\ref{e:uncertainty_pt_depth}), we can obtain the relation between the ground-truth point ${}^L \mathbf{p}_i^{gt}$ and its measurement ${}^L \mathbf{p}_i$
\begin{equation}
\label{e:uncertainty_pt_local}
    \begin{aligned}
        {}^L \mathbf{p}_i^{gt} 
        &= d^{gt}_i  \bm{\phi}_i^{gt} = \left( d_i + \delta_{d_i} \right)   \left( \bm{\phi}_i \boxplus_{\mathbb{S}^2} \delta_{\bm{\phi}_i} \right) \\
        &\approx \left( d_i + \delta_{d_i} \right) \left( \bm{\phi}_i + \lfloor \mathbf{N} \left(\bm{\phi}_i \right) \delta_{\bm{\phi}_i} \rfloor_{\wedge} \bm{\phi}_i \right) \\
        &= d_i \bm{\phi}_i + \bm{\phi}_i \delta_{d_i} + d_i \lfloor \mathbf{N} \left(\bm{\phi}_i \right) \delta_{\bm{\phi}_i} \rfloor_{\wedge} \bm{\phi}_i + \mathcal{O}\left(\delta_{d_i}\delta_{\bm{\phi}_i}\right) \\
        &\approx d_i \bm{\phi}_i + \bm{\phi}_i \delta_{d_i} - d_i \lfloor\bm{\phi}_i \rfloor_{\wedge} \mathbf{N} \left(\bm{\phi}_i \right) \delta_{\bm{\phi}_i} \\
        &= {}^L \mathbf{p}_i + \delta_{{}^L \mathbf{p}_i}        
    \end{aligned}
\end{equation}
where ${}^L \mathbf{p}_i \triangleq d_i \bm{\phi}_i$, and $\delta_{{}^L \mathbf{p}_i} \triangleq \bm{\phi}_i \delta_{d_i} - d_i \lfloor\bm{\phi}_i \rfloor_{\wedge} \mathbf{N} \left(\bm{\phi}_i \right) \delta_{\bm{\phi}_i} $. To get the formulation in~\cite{yuan2022efficient, yuan2021pixel}, rewrite $\delta_{{}^L \mathbf{p}_i}$ as shown
\begin{equation}
\label{e:uncertainty_pt_local_noise}
    \begin{aligned}
        \delta_{{}^L \mathbf{p}_i} 
        &= \bm{\phi}_i \delta_{d_i} - d_i \lfloor\bm{\phi}_i \rfloor_{\wedge} \mathbf{N} \left(\bm{\phi}_i \right) \delta_{\bm{\phi}_i} \\
        &= \begin{bmatrix} \bm{\phi}_i &  - d_i \lfloor\bm{\phi}_i \rfloor_{\wedge} \mathbf{N} \left(\bm{\phi}_i \right)\end{bmatrix} 
        \begin{bmatrix}
        \delta_{d_i} \\ \delta_{\bm{\phi}_i}    
        \end{bmatrix}.
    \end{aligned}
\end{equation}
We can let $ \mathbf{A}_i = \begin{bmatrix} \bm{\phi}_i &  - d_i \lfloor\bm{\phi}_i \rfloor_{\wedge} \mathbf{N} \left(\bm{\phi}_i \right)\end{bmatrix} \in \mathbb{R}_{3\times3}$, thus
\begin{equation}
\label{e:uncertainty_pt_local_covariance}
    \begin{aligned}
        \delta_{{}^L \mathbf{p}_i} 
        &\sim \mathcal{N}\left(\mathbf{0}, \bm{\Sigma}_{ {}^L \mathbf{p}_i} \right), \\
        \bm{\Sigma}_{ {}^L \mathbf{p}_i} 
        &= \mathbf{A}_i \begin{bmatrix} \Sigma_{d_i} & \mathbf{0}_{1\times2} \\ \mathbf{0}_{2\times1} & \bm{\Sigma}_{\phi_i} \end{bmatrix} \mathbf{A}_i^T \\
        &= \begin{bmatrix} \bm{\phi}_i &  - d_i \lfloor\bm{\phi}_i \rfloor_{\wedge} \mathbf{N} \left(\bm{\phi}_i \right)\end{bmatrix}
        \begin{bmatrix} \Sigma_{d_i} & \mathbf{0}_{1\times2} \\ \mathbf{0}_{2\times1} & \bm{\Sigma}_{\phi_i} \end{bmatrix}
        \begin{bmatrix} \bm{\phi}_i^T \\  - d_i (\lfloor\bm{\phi}_i \rfloor_{\wedge} \mathbf{N} \left(\bm{\phi}_i \right))^T\end{bmatrix}\\
        &= \bm{\phi}_i\Sigma_{d_i}\bm{\phi}_i^T + d_i^2\lfloor\bm{\phi}_i \rfloor_{\wedge} \mathbf{N} \left(\bm{\phi}_i \right) \bm{\Sigma}_{\phi_i} \mathbf{N} \left(\bm{\phi}_i \right)^T\lfloor\bm{\phi}_i \rfloor_{\wedge}^T.
    \end{aligned}
\end{equation}

Considering that the LiDAR point ${}^L \mathbf{p}_i$ will be further projected to the world frame through the estimated pose ${}^G_L\mathbf{T} = \left({}^G_L \mathbf{R}, {}^G_L \mathbf{t}\right) \in \rm{SE}\left(3\right)$, with estimation uncertainty $\left(\bm{\Sigma}_\mathbf{R}, \bm{\Sigma}_\mathbf{t}\right)$, by the following rigid transformation
\begin{equation}
\label{e:rigid_trans}
    {}^G\mathbf{p}_i = {}^G_L \mathbf{R} {}^L \mathbf{p}_i +  {}^G_L \mathbf{t}.
\end{equation}
Therefore, the uncertainty of the LiDAR points ${}^G \mathbf{p}_i$ can be obtained by
\begin{equation}
\label{e:uncertainty_pt_global_covariance}
    \begin{aligned}
        {}^G\mathbf{p}_i^{gt} 
        &= {}^G_L \mathbf{R}^{gt} {}^L \mathbf{p}_i^{gt} +  {}^G_L \mathbf{t}^{gt} \\
        &= {}^G_L \mathbf{R} \rm{Exp}\left( \delta_{\bm{\theta}} \right) \left({}^L \mathbf{p}_i + \delta_{{}^L \mathbf{p}_i} \right) + {}^G_L \mathbf{t} + \delta_{\mathbf{t}} \\
        &\approx {}^G_L \mathbf{R} \left(I + \lfloor \delta_{\bm{\theta}}\rfloor_{\wedge}\right) \left({}^L \mathbf{p}_i + \delta_{{}^L \mathbf{p}_i} \right) + {}^G_L \mathbf{t} + \delta_{\mathbf{t}} \\
        &= {}^G_L \mathbf{R}{}^L \mathbf{p}_i + {}^G_L \mathbf{R}\delta_{{}^L \mathbf{p}_i} + {}^G_L \mathbf{R}\lfloor \delta_{\bm{\theta}}\rfloor_{\wedge}{}^L \mathbf{p}_i + \mathcal{O}\left(\delta_{\bm{\theta}}, \delta_{{}^L \mathbf{p}_i}\right) + {}^G_L \mathbf{t} + \delta_{\mathbf{t}} \\
        &\approx {}^G\mathbf{p}_i +  {}^G_L \mathbf{R}\delta_{{}^L \mathbf{p}_i} - {}^G_L \mathbf{R}\lfloor {}^L\mathbf{p}_i\rfloor_{\wedge} \delta_{\bm{\theta}} + \delta_{\mathbf{t}}, \\
        \delta_{{}^G \mathbf{p}_i} 
        &= {}^G_L \mathbf{R}\delta_{{}^L \mathbf{p}_i} - {}^G_L \mathbf{R}\lfloor {}^L\mathbf{p}_i\rfloor_{\wedge} \delta_{\bm{\theta}} + \delta_{\mathbf{t}} \sim \mathcal{N}\left(\mathbf{0}, \bm{\Sigma}_{{}^G \mathbf{p}_i}\right), \\
        \bm{\Sigma}_{{}^G \mathbf{p}_i} 
        &= \bm{\Sigma}_{\mathbf{t}} + {}^G_L\mathbf{R} \bm{\Sigma}_{{}^L \mathbf{p}_i} {}^G_L\mathbf{R}^T + {}^G_L \mathbf{R}\lfloor {}^L\mathbf{p}_i\rfloor_{\wedge} \bm{\Sigma}_{\mathbf{R}} \left({}^G_L \mathbf{R}\lfloor {}^L\mathbf{p}_i\rfloor_{\wedge}\right)^T \\
        &= {}^G_L\mathbf{R} \bm{\Sigma}_{{}^L \mathbf{p}_i} {}^G_L\mathbf{R}^T + {}^G_L \mathbf{R}\lfloor {}^L\mathbf{p}_i\rfloor_{\wedge} \bm{\Sigma}_{\mathbf{R}} \lfloor {}^L\mathbf{p}_i\rfloor_{\wedge}^T {}^G_L \mathbf{R}^T + \bm{\Sigma}_{\mathbf{t}}
    \end{aligned}
\end{equation}
where $\delta_{\bm{\theta}} \in \mathbb{R}_3$ and $\delta_{\mathbf{t}} \in \mathbb{R}_3$ denote the small perturbations of the rotation and translation components of the estimated pose, respectively.

\subsubsection{Uncertainty of plane}
Referring to~\cite{liu2021balm, yuan2022efficient}, let a plane feature consist of a group of LiDAR points ${}^G\mathbf{p}_i \left(i=1,\dots,N\right)$, each has an uncertainty $\bm{\Sigma}_{{}^G \mathbf{p}_i} $ due to the measurement noise and pose estimation error as shown in (\ref{e:uncertainty_pt_global_covariance}). 
Denote the centroid and covariance matrix of the points by $\bar{\mathbf{p}}$ and $\mathbf{A}$, respectively
\begin{equation}
\label{e:plane_model}
    \bar{\mathbf{p}} = \frac{1}{N} \sum_{i=1}^{N} {}^G \mathbf{p}_i, \ \ \ \ \ \mathbf{A} = \frac{1}{N} \sum_{i=1}^N \left({}^G \mathbf{p}_i - \bar{\mathbf{p}} \right) \left({}^G \mathbf{p}_i - \bar{\mathbf{p}} \right)^T.
\end{equation}
Then, the plane can be represented by its normal vector $\mathbf{n}$, which is the eigenvector associated with the minimum eigenvalue of $\mathbf{A}$, and the point $\mathbf{q} =\bar{\mathbf{p}}$, which lies in this plane. 
As both $\mathbf{A}$ and $\mathbf{q}$ are dependent on ${}^G\mathbf{p}_i$, denote the plane parameters $\left(\mathbf{n},\mathbf{q}\right)$ as functions of ${}^G\mathbf{p}_i$
\begin{equation}
\label{e:plane_of_world_pt}
    \begin{bmatrix}\mathbf{n},\mathbf{q} \end{bmatrix}^T = \mathbf{f}\left({}^G\mathbf{p}_1, {}^G\mathbf{p}_2, \dots, {}^G\mathbf{p}_N \right).
\end{equation}

\begin{figure*}[htbp] 
    \centering
    \includegraphics[width=0.8\textwidth]{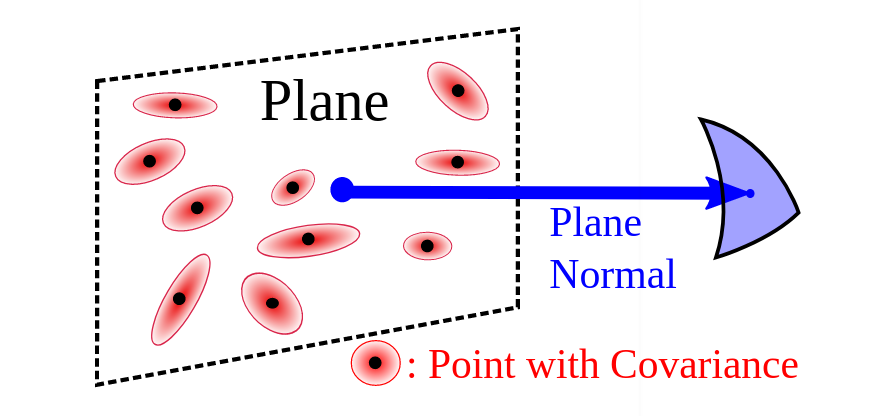}
    \caption{\textbf{The Uncertainty model of the plane normal.}}
    \label{fig:uncertainty_plane}
\end{figure*}

Next, we derive the uncertainty of plane shown in Figure~\ref{fig:uncertainty_plane} based on the uncertainty of point ${}^G\mathbf{p}_i$ in (\ref{e:uncertainty_pt_global_covariance}). 
The ground-truth normal vector $\mathbf{n}^{gt}$ and ground-truth position $\mathbf{q}^{gt}$ are
\begin{equation}
\label{e:plane_model_taylor}
    \begin{aligned}
        \begin{bmatrix}\mathbf{n}^{gt},\mathbf{q}^{gt} \end{bmatrix}^T 
        &= \mathbf{f}\left({}^G\mathbf{p}_1 + \delta_{{}^G\mathbf{p}_1}, {}^G\mathbf{p}_2+\delta_{{}^G\mathbf{p}_2}, \dots, {}^G\mathbf{p}_N + \delta_{{}^G\mathbf{p}_N}\right) \\
        &\approx \begin{bmatrix}\mathbf{n},\mathbf{q} \end{bmatrix}^T + \sum_{i=1}^N\frac{\partial \mathbf{f}}{\partial {}^G\mathbf{p}_i} \delta_{{}^G\mathbf{p}_i}, \\
        \delta_{\mathbf{n},\mathbf{q}} 
        &= \sum_{i=1}^N\frac{\partial \mathbf{f}}{\partial {}^G\mathbf{p}_i} \delta_{{}^G\mathbf{p}_i} \sim \mathcal{N}(\mathbf{0}, \bm{\Sigma}_{\mathbf{n},\mathbf{q}}) 
    \end{aligned}
\end{equation}
where $\frac{\partial \mathbf{f}}{\partial {}^G\mathbf{p}_i} = \begin{bmatrix} \frac{\partial \mathbf{n}}{\partial {}^G\mathbf{p}_i}, \frac{\partial \mathbf{q}}{\partial {}^G\mathbf{p}_i} \end{bmatrix}^T$.
Thus, we can get the uncertainty of the plane $\bm{\Sigma}_{\mathbf{n},\mathbf{q}} = \sum_{i=1}^N\frac{\partial \mathbf{f}}{\partial {}^G\mathbf{p}_i} \bm{\Sigma}_{{}^G\mathbf{p}_i}\frac{\partial \mathbf{f}}{\partial {}^G\mathbf{p}_i}^T$.

According to (\ref{e:plane_model}), we can easily take the derivative of $\mathbf{q}$ with respect to each point ${}^G\mathbf{p}_i$
\begin{equation}
\label{e:q_jacobian}
    \frac{\partial \mathbf{q}}{\partial {}^G\mathbf{p}_i} = \frac{1}{N}\mathbf{I}_3.
\end{equation}

Then, do the the derivative of $\mathbf{n}$ with respect to each point ${}^G\mathbf{p}_i$. The normal vector $\mathbf{n}$ is the eigenvector associated with the minimum eigenvalue $\lambda_3$ of $\mathbf{A}$ mentioned in (\ref{e:plane_model}).
Also, we have the definition of eigenvector and eigenvalue 
\begin{equation}
\label{e:eigenvector_eigenvalue}
    \begin{aligned}
        \mathbf{U} 
        &= \begin{bmatrix} \mathbf{u}_1 &  \mathbf{u}_2 &  \mathbf{u}_3 \end{bmatrix}, \ \ \ 
        \bm{\Lambda} = \begin{bmatrix} \lambda_1 & 0 & 0 \\ 0 & \lambda_2 & 0 \\ 0 & 0 & \lambda_3 \end{bmatrix}, \ \ \ \mathbf{U}^T \mathbf{U} = \mathbf{I}, \\
        \mathbf{A} 
        &= \mathbf{U} \bm{\Lambda} \mathbf{U}^T, \ \ \
        \bm{\Lambda} = \mathbf{U}^T \mathbf{A} \mathbf{U}, \ \ \ 
        \mathbf{A} \mathbf{U} = \mathbf{U} \bm{\Lambda},  \ \ \ 
        \mathbf{U}^T \mathbf{A} = \bm{\Lambda} \mathbf{U}^T.
    \end{aligned}
\end{equation}
Take the derivative of $\mathbf{U}^T\mathbf{U}$ with respect to each point ${}^G\mathbf{p}_i$
\begin{equation}
\label{e:UU_p_deriv}
    \begin{aligned}
        & \mathbf{U}^T \frac{\partial \mathbf{U}}{\partial {}^G\mathbf{p}_i} + \frac{\partial \mathbf{U}}{\partial {}^G\mathbf{p}_i}^T \mathbf{U} = \mathbf{0 }, \\
        & \rm{Let} \ \ \mathbf{C}^{\mathbf{p}_i}  = \mathbf{U}^T \frac{\partial \mathbf{U}}{\partial {}^G\mathbf{p}_i}, \\ 
        & \to \ \  \mathbf{C}^{\mathbf{p}_i} + {\mathbf{C}^{\mathbf{p}_i}}^T = \mathbf{0} 
    \end{aligned}
\end{equation}
where $\mathbf{C}^{\mathbf{p}_i}$ is a skew-symmetric matrix whose diagonal elements are zeros. 
Using (\ref{e:UU_p_deriv}), take the derivative of $\bm{\Lambda}$ with respect to each point ${}^G\mathbf{p}_i$
\begin{equation}
\label{e:eigenvalue_p_deriv}
    \begin{aligned}
        \frac{\partial \bm{\Lambda}}{\partial {}^G\mathbf{p}_i} 
        &=  \frac{\partial \mathbf{U}^T}{\partial {}^G\mathbf{p}_i} \mathbf{A} \mathbf{U} + \mathbf{U}^T  \frac{\partial \mathbf{A}}{\partial {}^G\mathbf{p}_i}  \mathbf{U} +   \mathbf{U}^T \mathbf{A} \frac{\partial \mathbf{U}}{\partial {}^G\mathbf{p}_i} \\
        &= \mathbf{U}^T  \frac{\partial \mathbf{A}}{\partial {}^G\mathbf{p}_i}  \mathbf{U} +  \bm{\Lambda} \mathbf{U}^T  \frac{\partial \mathbf{U}}{\partial {}^G\mathbf{p}_i} + \frac{\partial \mathbf{U}^T}{\partial {}^G\mathbf{p}_i} \mathbf{U} \bm{\Lambda}\\
        &= \mathbf{U}^T  \frac{\partial \mathbf{A}}{\partial {}^G\mathbf{p}_i}  \mathbf{U} + \bm{\Lambda} \mathbf{C}^{\mathbf{p}_i} - \mathbf{C}^{\mathbf{p}_i} \bm{\Lambda}.
    \end{aligned}
\end{equation}
Since $\bm{\Lambda}$ is diagonal, $\frac{\partial \bm{\Lambda}}{\partial {}^G\mathbf{p}_i}$, for off-diagonal element in (\ref{e:eigenvalue_p_deriv}), we have
\begin{equation}
\label{e:offdiagnal_element}
    \begin{aligned}
        0 
        &= \mathbf{u}_m^T \frac{\partial \mathbf{A}}{\partial {}^G\mathbf{p}_i} \mathbf{u}_n + \lambda_m \mathbf{C}^{\mathbf{p}_i}_{m,n} -  \mathbf{C}^{\mathbf{p}_i}_{m,n} \lambda_n,   \\
        &\to  \mathbf{u}_m^T \frac{\partial \mathbf{A}}{\partial {}^G\mathbf{p}_i} \mathbf{u}_n 
        = \left(\lambda_n - \lambda_m\right) \mathbf{C}^{\mathbf{p}_i}_{m,n}, \\
        &\to \mathbf{C}^{\mathbf{p}_i}_{m,n} 
        = \frac{1}{\lambda_n -\lambda_m}  \mathbf{u}_m^T\frac{\partial \mathbf{A}}{\partial {}^G\mathbf{p}_i} \mathbf{u}_n \ \ , m \neq n
    \end{aligned}
\end{equation}
where $\mathbf{C}^{\mathbf{p}_i}_{m,n} $ is the $m$-th row and $n$-th column element in $\mathbf{C}^{\mathbf{p}_i}$. 
Using (\ref{e:UU_p_deriv}) and (\ref{e:offdiagnal_element}), gives
\begin{equation}
\label{e:cq_element}
    \mathbf{C}^{\mathbf{p}_i}_{m,n} = \left\{
    \begin{aligned}
        \frac{1}{\lambda_n -\lambda_m} & \mathbf{u}_m^T\frac{\partial \mathbf{A}}{\partial {}^G\mathbf{p}_i} \mathbf{u}_n  &, \ \ m \neq n \\
         & \mathbf{0}_{1\times3} \  &, \ \ m= n
    \end{aligned}
    \right. 
\end{equation}
According to the formulation of $\mathbf{C}^{\mathbf{p}_i}$ in (\ref{e:UU_p_deriv}), yield 
\begin{equation}
\label{e:U_p_deriv}
    \frac{\partial \mathbf{U}}{\partial {}^G\mathbf{p}_i} = \mathbf{U} \mathbf{C}^{\mathbf{p}_i} = \mathbf{U}\mathbf{U}^T\frac{\partial \mathbf{U}}{\partial {}^G\mathbf{p}_i}.
\end{equation}
Using $\mathbf{e}_k$, a $3 \times 1$ vector in which the $k$-th element is $1$ and the rests $0$, combined with (\ref{e:U_p_deriv}), gives
\begin{equation}
\label{e:uk_p_deriv}
    \begin{aligned}
        \frac{\partial \mathbf{u}_k}{\partial {}^G\mathbf{p}_i} 
        &= \frac{\partial \mathbf{U}}{\partial {}^G\mathbf{p}_i}  \mathbf{e}_k, \\
        \rm{Let} {}^G\mathbf{p}_i
        &= \begin{bmatrix} x_i & y_i & z_i \end{bmatrix}, \\
        \to \frac{\partial \mathbf{u}_k}{\partial {}^G\mathbf{p}_i} 
        &= \begin{bmatrix} \frac{\partial \mathbf{U}}{\partial x_i}  \mathbf{e}_k & \frac{\partial \mathbf{U}}{\partial y_i}  \mathbf{e}_k & \frac{\partial \mathbf{U}}{\partial z_i}  \mathbf{e}_k \end{bmatrix} \\
        &= \begin{bmatrix} \mathbf{U} \mathbf{C}^{x_i}  \mathbf{e}_k & \mathbf{U} \mathbf{C}^{y_i}  \mathbf{e}_k & \mathbf{U} \mathbf{C}^{z_i}  \mathbf{e}_k \end{bmatrix} \\
        &= \mathbf{U} \begin{bmatrix} \mathbf{C}^{x_i}  \mathbf{e}_k & \mathbf{C}^{y_i}  \mathbf{e}_k &  \mathbf{C}^{z_i}  \mathbf{e}_k \end{bmatrix} \\
        &= \mathbf{U} 
        \begin{bmatrix} 
            \mathbf{C}^{x_i}_{1,k}  & \mathbf{C}^{y_i}_{1,k} &  \mathbf{C}^{z_i}_{1,k} \\
            \mathbf{C}^{x_i}_{2,k}  & \mathbf{C}^{y_i}_{2,k} &  \mathbf{C}^{z_i}_{2,k} \\
            \mathbf{C}^{x_i}_{3,k}  & \mathbf{C}^{y_i}_{3,k} &  \mathbf{C}^{z_i}_{3,k} 
        \end{bmatrix}
    \end{aligned}
\end{equation}
where $\mathbf{C}^{x_i} \triangleq \mathbf{U}^T \frac{\partial \mathbf{U}}{\partial x_i} \in \mathbb{R}_{3\times3}$, $\mathbf{C}^{y_i} \triangleq \mathbf{U}^T \frac{\partial \mathbf{U}}{\partial y_i} \in \mathbb{R}_{3\times3}$ and $\mathbf{C}^{z_i} \triangleq \mathbf{U}^T \frac{\partial \mathbf{U}}{\partial z_i} \in \mathbb{R}_{3\times3}$. 
To simplify (\ref{e:uk_p_deriv}), define $\mathbf{F}^{\mathbf{p}_i}_{m,n}=\begin{bmatrix}\mathbf{C}^{x_i}_{m,n} & \mathbf{C}^{y_i}_{m,n} & \mathbf{C}^{z_i}_{m,n}\end{bmatrix} \in \mathbb{R}_{1\times3}, \ \ m,n \in \{1,2,3\}$
\begin{equation}
\label{e:uk_p_deriv_simple}
    \frac{\partial \mathbf{u}_k}{\partial {}^G\mathbf{p}_i} 
    = \mathbf{U} \begin{bmatrix}\mathbf{F}^{\mathbf{p}_i}_{1,k} \\\mathbf{F}^{\mathbf{p}_i}_{2,k} \\\mathbf{F}^{\mathbf{p}_i}_{3,k}  \end{bmatrix} = \mathbf{U} \mathbf{F}^{\mathbf{p}_i}_k.
\end{equation}
So far, only $\mathbf{F}^{\mathbf{p}_i}_{m,n}$ has not been fully derived. According to (\ref{e:cq_element}), hence
\begin{equation}
\label{e:fq}
    \mathbf{F}^{\mathbf{p}_i}_{m,n} = \left\{
    \begin{aligned}
        \frac{1}{\lambda_n -\lambda_m} &\frac{\partial  \mathbf{u}_m^T\mathbf{A} \mathbf{u}_n}{\partial {}^G\mathbf{p}_i}  &, \ \ m \neq n \\
         & \mathbf{0}_{1\times3} \  &, \ \ m= n
    \end{aligned}
    \right.  
\end{equation}
To further obtain the specific form of $\mathbf{F}^{\mathbf{p}_i}_{m,n}$, derive $\frac{\partial  \mathbf{u}_m^T\mathbf{A} \mathbf{u}_n}{\partial {}^G\mathbf{p}_i}$ ($\mathbf{u}_m$ and $\mathbf{u}_n$ are viewed as constant vector)
\begin{equation}
\label{e:umAun_p_deriv}
    \begin{aligned}
        \frac{\partial  \mathbf{u}_m^T\mathbf{A} \mathbf{u}_n}{\partial {}^G\mathbf{p}_i}
        &= \frac{1}{N}\sum_{j=1}^N \frac{\partial  \mathbf{u}_m^T \left({}^G \mathbf{p}_j - \bar{\mathbf{p}} \right) \left({}^G \mathbf{p}_j - \bar{\mathbf{p}} \right)^T \mathbf{u}_n}{\partial {}^G\mathbf{p}_i} \\
        &= \frac{1}{N}\sum_{j=1}^N \frac{\partial  \mathbf{u}_m^T \left({}^G \mathbf{p}_j - \bar{\mathbf{p}} \right) \mathbf{u}_n^T\left({}^G \mathbf{p}_j - \bar{\mathbf{p}} \right)}{\partial {}^G\mathbf{p}_i}, \\
        \rm{Let} \ \ \mathbf{e}_j 
        &= \mathbf{u}_m^T \left({}^G \mathbf{p}_j - \bar{\mathbf{p}} \right) \in \mathbb{R},  \mathbf{f}_j = \mathbf{u}_n^T \left({}^G \mathbf{p}_j - \bar{\mathbf{p}} \right) \in \mathbb{R}, \\
        \to \frac{\partial  \mathbf{u}_m^T\mathbf{A} \mathbf{u}_n}{\partial {}^G\mathbf{p}_i} 
        &= \frac{1}{N}\sum_{j=1}^N \frac{\partial \mathbf{e}_j\mathbf{f}_j}{\partial {}^G\mathbf{p}_i} 
        = \frac{1}{N}\sum_{j=1}^N \mathbf{e}_j\frac{\partial \mathbf{f}_j}{\partial {}^G\mathbf{p}_i} + \mathbf{f}_j \frac{\partial \mathbf{e}_j}{\partial {}^G\mathbf{p}_i}.
    \end{aligned}
\end{equation}
Then, 
\begin{equation}
\label{e:umAun_p_deriv_ef}
    \begin{aligned}
        \sum_{j=1}^N \mathbf{e}_j\frac{\partial \mathbf{f}_j}{\partial {}^G\mathbf{p}_i} 
        &= \sum_{j=1}^N \mathbf{u}_m^T \left({}^G \mathbf{p}_j - \bar{\mathbf{p}} \right) \frac{\partial \mathbf{u}_n^T \left({}^G \mathbf{p}_j - \bar{\mathbf{p}} \right)}{\partial {}^G\mathbf{p}_i}\\
        &= \mathbf{u}_m^T \left({}^G \mathbf{p}_i - \bar{\mathbf{p}} \right) \frac{\partial \mathbf{u}_n^T \left({}^G \mathbf{p}_i - \bar{\mathbf{p}} \right)}{\partial {}^G\mathbf{p}_i} 
        + \sum_{j=1,j\neq i}^N \mathbf{u}_m^T \left({}^G \mathbf{p}_j - \bar{\mathbf{p}} \right) \frac{\partial \mathbf{u}_n^T \left({}^G \mathbf{p}_j - \bar{\mathbf{p}} \right)}{\partial {}^G\mathbf{p}_i} \\
        &= \left({}^G \mathbf{p}_i - \bar{\mathbf{p}}\right)^T \mathbf{u}_m \mathbf{u}_n^T (1 - \frac{1}{N}) + \sum_{j=1,j\neq i}^N \left({}^G \mathbf{p}_j - \bar{\mathbf{p}}\right)^T \mathbf{u}_m \mathbf{u}_n^T (- \frac{1}{N}) \\
        & = \left({}^G \mathbf{p}_i - \bar{\mathbf{p}}\right)^T \mathbf{u}_m \mathbf{u}_n^T (1 - \frac{1}{N}) + \left({}^G \mathbf{p}_i - \bar{\mathbf{p}}\right)^T \mathbf{u}_m \mathbf{u}_n^T  \frac{1}{N} \\
        &= \left({}^G \mathbf{p}_i - \bar{\mathbf{p}}\right)^T \mathbf{u}_m \mathbf{u}_n^T.
    \end{aligned}
\end{equation}
Similarly, derive that $\sum_{j=1}^N \mathbf{f}_j  \frac{\partial \mathbf{e}_j}{\partial {}^G\mathbf{p}_i} =  \left({}^G \mathbf{p}_i - \bar{\mathbf{p}}\right)^T \mathbf{u}_n \mathbf{u}_m^T$. 
Using (\ref{e:umAun_p_deriv}) and (\ref{e:umAun_p_deriv_ef}), we can directly give $\frac{\partial  \mathbf{u}_m^T\mathbf{A} \mathbf{u}_n}{\partial {}^G\mathbf{p}_i} = \left({}^G \mathbf{p}_i - \bar{\mathbf{p}}\right)^T \left(\mathbf{u}_m \mathbf{u}_n^T + \mathbf{u}_n \mathbf{u}_m^T \right)$ when $ m \neq n$. Also, 
\begin{equation}
\label{e:fq_final}
    \mathbf{F}^{\mathbf{p}_i}_{m,n} = \left\{
    \begin{aligned}
        \frac{\left({}^G \mathbf{p}_i - \bar{\mathbf{p}}\right)^T}{N\left(\lambda_n -\lambda_m\right)} &\left(\mathbf{u}_m \mathbf{u}_n^T + \mathbf{u}_n \mathbf{u}_m^T \right)  &, \ \ m \neq n \\
         & \mathbf{0}_{1\times3} \  &, \ \ m= n
    \end{aligned}
    \right.      
\end{equation}
Refer to (\ref{e:uk_p_deriv_simple}) and (\ref{e:fq_final}), recall that the normal vector $\mathbf{n}$ is the eigenvector associated with the minimum eigenvalue $\lambda_3$ of $\mathbf{A}$
\begin{equation}
\label{e:normal_jacobian}
    \frac{\partial \mathbf{n}}{\partial {}^G\mathbf{p}_i} 
    = \mathbf{U} \begin{bmatrix}\mathbf{F}^{\mathbf{p}_i}_{1,3} \\\mathbf{F}^{\mathbf{p}_i}_{2,3} \\\mathbf{F}^{\mathbf{p}_i}_{3,3}  \end{bmatrix}, \ \ 
    \mathbf{F}^{\mathbf{p}_i}_{m,3} = \left\{
    \begin{aligned}
        \frac{\left({}^G \mathbf{p}_i - \bar{\mathbf{p}}\right)^T}{N\left(\lambda_3 -\lambda_m\right)} &\left(\mathbf{u}_m \mathbf{n}^T + \mathbf{n} \mathbf{u}_m^T \right)  &, \ \ m \neq 3 \\
         & \mathbf{0}_{1\times3} \  &, \ \ m= 3
    \end{aligned}
    \right.     
\end{equation}
Finally, give the uncertainty of plane 
\begin{equation}
\label{e:uncertainty_plane}
    \begin{aligned}
        \bm{\Sigma}_{\mathbf{n},\mathbf{q}} 
        &= \sum_{i=1}^N\frac{\partial \mathbf{f}}{\partial {}^G\mathbf{p}_i} \bm{\Sigma}_{{}^G\mathbf{p}_i}\frac{\partial \mathbf{f}}{\partial {}^G\mathbf{p}_i}^T \\
        &= \sum_{i=1}^N \begin{bmatrix} \frac{\partial \mathbf{n}}{\partial {}^G\mathbf{p}_i} \\ \frac{\partial \mathbf{q}}{\partial {}^G\mathbf{p}_i} \end{bmatrix} 
        \bm{\Sigma}_{{}^G\mathbf{p}_i} 
        \begin{bmatrix} \frac{\partial \mathbf{n}}{\partial {}^G\mathbf{p}_i}^T & \frac{\partial \mathbf{q}}{\partial {}^G\mathbf{p}_i}^T \end{bmatrix} \\
        &= \sum_{i=1}^N  \begin{bmatrix} 
            \frac{\partial \mathbf{n}}{\partial {}^G\mathbf{p}_i} \bm{\Sigma}_{{}^G\mathbf{p}_i} \frac{\partial \mathbf{n}}{\partial {}^G\mathbf{p}_i}^T & 
            \frac{\partial \mathbf{n}}{\partial {}^G\mathbf{p}_i} \bm{\Sigma}_{{}^G\mathbf{p}_i} \frac{\partial \mathbf{q}}{\partial {}^G\mathbf{p}_i}^T \\
            \frac{\partial \mathbf{q}}{\partial {}^G\mathbf{p}_i} \bm{\Sigma}_{{}^G\mathbf{p}_i} \frac{\partial \mathbf{n}}{\partial {}^G\mathbf{p}_i}^T &
            \frac{\partial \mathbf{q}}{\partial {}^G\mathbf{p}_i} \bm{\Sigma}_{{}^G\mathbf{p}_i} \frac{\partial \mathbf{q}}{\partial {}^G\mathbf{p}_i}^T
            \end{bmatrix}.
    \end{aligned}
\end{equation}

\subsection{Map Construction and Update}
\label{sec:voxel_const}
Referring to~\cite{yuan2022efficient}, a coarse-to-fine voxel mapping method was proposed, which can build a rough voxel map when the point cloud is sparse and refine the map when more points are received. 
Since LiDAR points are acquired sequentially, each scan is initially sparse, especially in large-scale outdoor environments. 
Under such conditions, conventional fine-to-coarse surfel mapping often yields too few reliable planes, motivating the coarse-to-fine VoxelMap strategy to improve plane construction from sparse observations while reducing reliance on long accumulation intervals.

To construct the VoxelMap in a coarse-to-fine manner, the global space is first partitioned into coarse voxels, which are indexed by a hash table. Each occupied voxel is further represented by an octree, where plane fitting is recursively performed on the contained points. If the points within a voxel satisfy a planar criterion, the plane parameters and their uncertainty are estimated and stored; otherwise, the voxel is subdivided into eight child voxels until a valid plane is found or the maximum octree depth is reached. In this way, voxels of different sizes are adaptively generated, each maintaining a single probabilistic plane feature fitted from its raw LiDAR points.

For online operation, newly registered LiDAR points are continuously inserted into the VoxelMap according to the estimated pose as shown in Figure~\ref{fig:system_pipeline}. If a point falls into an unoccupied region, a new voxel is created; otherwise, the corresponding plane parameters and uncertainty are updated. To limit the computational growth caused by accumulating historical points, the update process exploits the empirical convergence of plane uncertainty. Once the plane estimate becomes sufficiently stable, historical points are discarded and only the plane parameters, covariance, and a small set of recent points are retained. If the normal vector estimated from newly arriving points deviates significantly from the stored plane, the voxel is regarded as changed and is reconstructed.

\section{Error-State Kalman Filter for LiDAR-Inertial Odometry}
\label{sec:lio}
This section presents the tightly-coupled LiDAR-Inertial Odometry formulation under the error-state Kalman filter framework. 
Following the system model introduced in Section~\ref{sec:preliminaries} and the probabilistic VoxelMap representation established in Section~\ref{sec:voxelmap}, the complete estimation pipeline is formulated, including IMU forward propagation, backward propagation for motion compensation, point-to-plane residual construction, and the iterated Kalman update.

To estimate the states in the state formulation (\ref{e:states_k}), use an iterated extended Kalman filter like~\cite{fastlio2}.
Assume the optimal state estimate of the last LiDAR scan at $t_{k-1}$ is $\bar{\mathbf{x}}_{k-1}$ with covariance matrix $\bar{\mathbf P}_{k-1}$. 
Then $\bar{\mathbf P}_{k-1}$ represents the covariance of the random error state vector defined below:
\begin{equation}
\label{e:error_state_def}
    \widetilde{\mathbf{x}}_{k-1} \doteq {\mathbf{x}}_{k-1} \boxminus \bar{\mathbf{x}}_{k-1} 
    = \begin{bmatrix} \delta \bm{\theta}^T & {}^G_I\widetilde{\mathbf{t}}^T & {}^G_I\widetilde{\mathbf{v}}^T & \widetilde{\mathbf{b}}_{\bm{\omega}}^T &  \widetilde{\mathbf{b}}_{\mathbf{a}}^T & {}^G\widetilde{\mathbf{g}}^T\end{bmatrix}^T
\end{equation}
where $\delta \bm{\theta}=\rm{Log}\left({}^G_I\bar{\mathbf{R}}^T {}^G_I \mathbf{R}\right)$ is the attitude error and the rest are standard additive errors (i.e., the error in the estimate $\bar{\mathbf{x}} $ of a quantity $\mathbf{x}$ is $\widetilde{\mathbf{x}} = \mathbf{x} - \bar{\mathbf{x}}$).
Intuitively, the attitude error $\delta \bm{\theta} $ describes the (small) deviation between the true and estimated attitude. 
The main advantage of this error definition is that it allows us to represent the uncertainty of attitude by the $3\times3$ covariance matrix $ \mathbb{E} \{\delta \bm{\theta} \delta\bm{\theta}^T\}$. 
Since the attitude has $3$ degree of freedom (DOF), this is a minimal representation.

\begin{figure*}[htbp]
    \centering
    \includegraphics[width=0.8\textwidth]{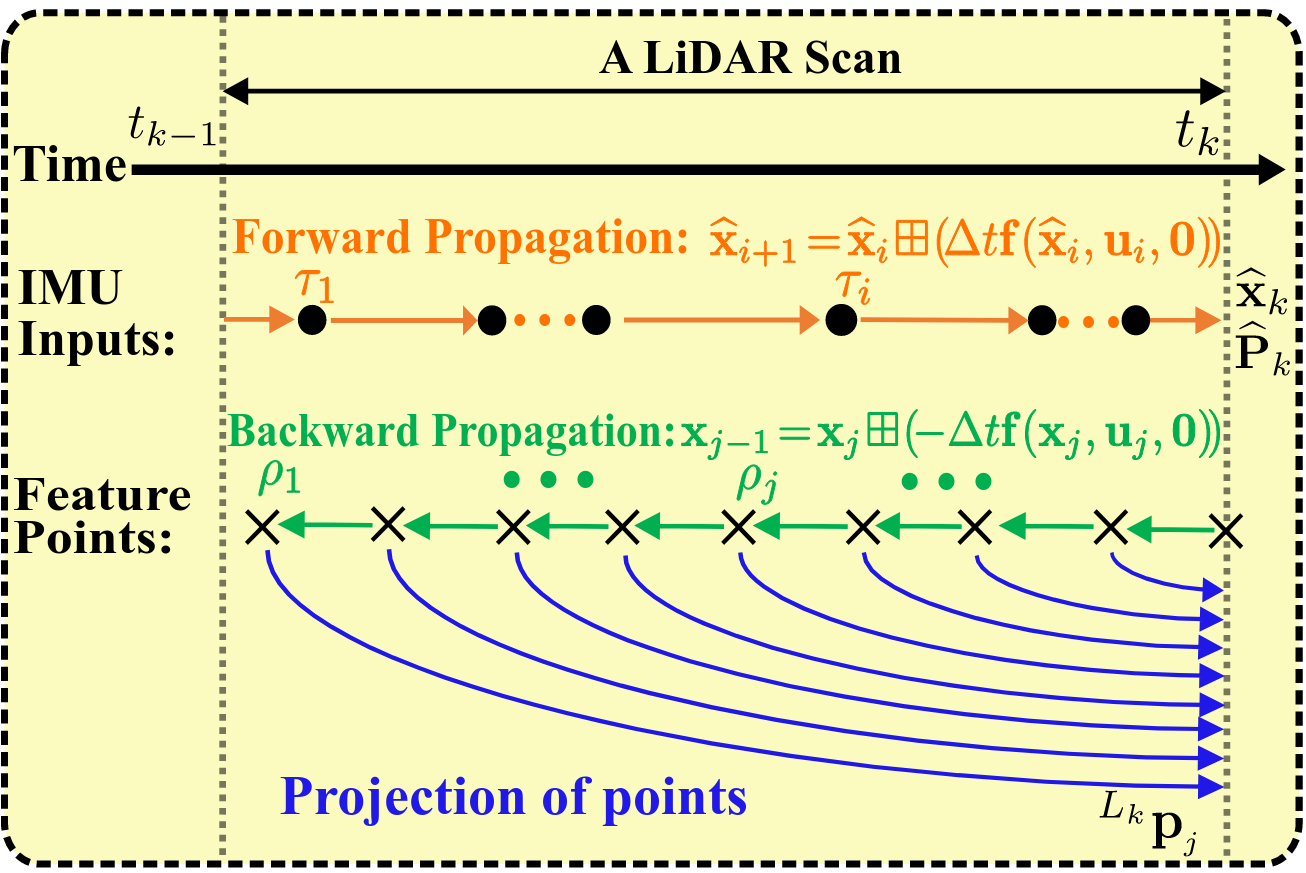}
    \caption{\textbf{The forward and backward propagation.} 
    }
    \label{fig:state_propagation}
\end{figure*}

\subsection{IMU Forward Propagation}
\label{sec:lio_forward}
The forward propagation is performed upon receiving each IMU measurement (see Figure~\ref{fig:state_propagation}).
Specifically, the state is propagated following (\ref{e:kine_model_discrete}) by setting the process noise $\mathbf{w}_i$ to zero
\begin{equation}
\label{e:forward_prop}
    \begin{aligned}
    \widehat{\mathbf{x}}_{i+1} 
    &= \widehat{\mathbf{x}}_i \boxplus \left( \Delta t \mathbf{f} \left(\widehat{\mathbf{x}}_i, \mathbf{u}_i, \mathbf{0}\right) \right); \ \widehat{\mathbf x}_{0} = \bar{\mathbf{x}}_{k-1}
    \end{aligned}
\end{equation}
where $\Delta t = \tau_{i+1} - \tau_{i}$. 
To propagate the covariance, we use the error state dynamic model obtained below
\begin{equation}
\label{e:error_state_model}
    \begin{aligned}
        \widetilde{\mathbf{x}}_{i+1} 
        &= \mathbf{x}_{i+1} \boxminus \widehat{\mathbf{x}}_{i+1} \\
        &= \left( \mathbf{x}_i \boxplus \Delta t \mathbf{f}\left( \mathbf{x}_i, \mathbf{u}_i, \mathbf{w}_i \right) \right) \boxminus \left(\widehat{\mathbf{x}}_i \boxplus \Delta t \mathbf{f} \left( \widehat{\mathbf{x}}_i, \mathbf{u}_i, \mathbf{0} \right) \right) \\
        &\approx \mathbf{F}_{\widetilde{\mathbf{x}}} \widetilde{\mathbf{x}}_i + \mathbf{F}_{\mathbf{w}} \mathbf{w}_i
    \end{aligned}
\end{equation}
where the matrices $\mathbf{F}_{\widetilde{\mathbf{x}}}$ and $\mathbf{F}_{\mathbf{w}}$ are derived below.
Denoting the covariance of white noises $\mathbf{w}$ as $\mathbf{Q}$, then the propagated covariance $\widehat{\mathbf{P}}_i$ can be computed recursively as
\begin{equation}
\label{e:cov_forward}
    \begin{aligned}
        \widehat{\mathbf{P}}_{i+1} 
        &=\mathbf{F}_{\widetilde{\mathbf{x}}}\widehat{\mathbf{P}}_i\mathbf{F}_{\widetilde{\mathbf{x}}}^T + \mathbf{F}_{\mathbf{w}} \mathbf{Q} \mathbf{F}_{\mathbf{w}}^T; \ \widehat{\mathbf{P}}_0 = \bar{\mathbf{P}}_{k-1}.
    \end{aligned}
\end{equation}
The propagation continues until the end time of a new scan at $t_{k}$ where the propagated state and covariance are denoted as $\widehat{\mathbf{x}}_{k}, \widehat{\mathbf{P}}_k$.  
Thus, $\widehat{\mathbf{P}}_{k}$ represents the covariance of the propagation error $\mathbf{x}_k \boxminus \widehat{\mathbf{x}}_k$. 
Here, two equivalent derivations of $\mathbf{F}_{\widetilde{\mathbf{x}}}$ and $\mathbf{F}_{\mathbf{w}}$ are considered: (i) direct first-order Taylor linearization of the discrete propagation, and (ii) derivation of the continuous-time error dynamics $\dot{\widetilde{\mathbf{x}}}$ followed by discretization.

\subsubsection{Method 1: Direct Discrete-time Linearization}
First, consider a general discrete-time nonlinear system using first-order Taylor linearization
\begin{equation}
\label{e:nonlinear_sys}
    \begin{aligned}
        \mathbf{x}_{i} 
        &= f\left(\mathbf{x}_{i-1},\,\mathbf{u}_{i-1},\,\mathbf{w}_{i-1}\right), \\
        \to \widehat{\mathbf{x}}_i \boxplus \widetilde{\mathbf{x}}_i 
        &= f\left(\widehat{\mathbf{x}}_{i-1} \boxplus \widetilde{\mathbf{x}}_{i-1}, \mathbf{u}_{i-1}, \mathbf{w}_{i-1}\right) \\
        &\approx f\left(\widehat{\mathbf{x}}_{i-1}, \mathbf{u}_{i-1}, \mathbf{0}\right) + \mathbf{F}  \widetilde{\mathbf{x}}_{i-1} + \mathbf{G} \mathbf{w}_{i-1}
    \end{aligned}
\end{equation}
where $\mathbf{u}_{i-1}$ and $\mathbf{w}_{i-1}$ denote the input and process noise, $\mathbf{F} \triangleq \left.\frac{\partial f}{\partial \mathbf{x}_{i-1}}\right|_{(\widehat{\mathbf{x}}_{i-1}, \mathbf{u}_{i-1},\mathbf{0})}$ and
$\mathbf{G} \triangleq \left.\frac{\partial f}{\partial \mathbf{w}_{i-1}}\right|_{(\widehat{\mathbf{x}}_{i-1},\mathbf{u}_{i-1},\mathbf{0})}$.

Let $\mathbf{g}\left(\widetilde{\mathbf{x}}_i, \mathbf{w}_i\right) \triangleq  \Delta t \mathbf{f}\left( \mathbf{x}_i, \mathbf{u}_i, \mathbf{w}_i \right) = \Delta t \mathbf{f}\left(\widehat{\mathbf{x}}_i \boxplus \widetilde{\mathbf{x}}_i, \mathbf{u}_i, \mathbf{w}_i \right)$, thus (\ref{e:error_state_model}) can be rewritten as 
\begin{equation}
\label{e:error_state_model_g}
    \begin{aligned}
        \widetilde{\mathbf{x}}_{i+1} 
        &= \left( \mathbf{x}_i \boxplus \Delta t \mathbf{f}\left( \mathbf{x}_i, \mathbf{u}_i, \mathbf{w}_i \right) \right) \boxminus \left(\widehat{\mathbf{x}}_i \boxplus \Delta t \mathbf{f} \left( \widehat{\mathbf{x}}_i, \mathbf{u}_i, \mathbf{0} \right) \right) \\
        &=  \left( \left( \widehat{\mathbf{x}}_i \boxplus \widetilde{\mathbf{x}}_i \right) \boxplus \mathbf{g}\left(\widetilde{\mathbf{x}}_i, \mathbf{w}_i\right) \right) \boxminus \left(\widehat{\mathbf{x}}_i \boxplus \mathbf{g}\left(\mathbf{0}, \mathbf{0}\right) \right) \\
        &\triangleq \mathbf{G}\left(\widetilde{\mathbf{x}}_i, \mathbf{g}\left(\widetilde{\mathbf{x}}_i, \mathbf{w}_i\right)\right). 
    \end{aligned}
\end{equation}
Following (\ref{e:nonlinear_sys}), linearize the state model at $\widetilde{\mathbf{x}}_i = \mathbf{0}$ and $\mathbf{w}_i=\mathbf{0}$ 
\begin{equation}
\label{e:F_partial_xw}
    \begin{aligned}
        \mathbf{F}_{\widetilde{\mathbf{x}}} 
        &= \left.\frac{\partial  \mathbf{G}\left(\widetilde{\mathbf{x}}_i, \mathbf{g}\left(\mathbf{0}, \mathbf{0}\right)\right)}{\partial \widetilde{\mathbf{x}}_i} 
        + \frac{\partial  \mathbf{G}\left(\mathbf{0}, \mathbf{g}\left(\widetilde{\mathbf{x}}_i, \mathbf{0}\right)\right)}{\partial \mathbf{g}\left(\widetilde{\mathbf{x}}_i, \mathbf{0}\right)} \frac{\partial \mathbf{g}\left(\widetilde{\mathbf{x}}_i, \mathbf{0}\right)}{\partial \widetilde{\mathbf{x}}_i} \right|_{\widetilde{\mathbf{x}}_i = \mathbf{0}}, \\
        \mathbf{F}_{\mathbf{w}}
        &=  \left.\frac{\partial  \mathbf{G}\left(\mathbf{0}, \mathbf{g}\left(\mathbf{0}, \mathbf{w}_i\right)\right)}{\partial \mathbf{g}\left(\mathbf{0}, \mathbf{w}_i\right)} \frac{\partial \mathbf{g}\left(\mathbf{0}, \mathbf{w}_i\right)}{\partial \mathbf{w}_i} \right|_{\mathbf{w}_i =\mathbf{0}}.
    \end{aligned}
\end{equation}

For manifold-valued expressions of the form $\mathbf{E} = \left(\left(\mathbf{a}\boxplus \mathbf{b}\right)\boxplus \mathbf{c}\right)\boxminus \mathbf{d}$, the required Jacobians are evaluated using the differential rules of the $\boxplus/\boxminus$ calculus on manifolds 
\begin{equation}
\label{e:manifold_rules}
    \frac{\partial \mathbf{E}}{\partial \mathbf{b}} =  \frac{\partial \mathbf{E}}{\partial \mathbf{c}} = \mathbf{I}_{n\times n}, \ \  \mathbf{a,b,c,d}\in \mathbb{R}_n. 
\end{equation}
Specifically for $\mathbf{b,c}\in \mathbb{R}_3, \mathbf{a,d}\in \rm{SO}\left(3\right)$, the Jacobians are different. $\frac{\partial \mathbf{E}}{\partial \mathbf{b}}$ is derived as 
\begin{equation}
\label{e:manifold_rules_so3_b}
    \begin{aligned}
        \mathbf{E} 
        = \left(\left(\mathbf{a}\boxplus \mathbf{b}\right)\boxplus \mathbf{c}\right)\boxminus \mathbf{d}
        &= \rm{Log} \left( \mathbf{d}^T \left(\mathbf{a}\boxplus \mathbf{b}\right)\boxplus \mathbf{c}\right), \\
        \to \rm{Exp} \left( \mathbf{E} \right) 
        &= \mathbf{d}^T \mathbf{a} \ \rm{Exp} \left(\mathbf{b}\right) \ \rm{Exp} \left( \mathbf{c} \right), \\
        \to \rm{Exp} \left( \mathbf{E} + \delta \mathbf{E}\right) 
        &= \mathbf{d}^T \mathbf{a} \ \rm{Exp} \left(\mathbf{b}  + \delta \mathbf{b}\right) \ \rm{Exp} \left( \mathbf{c} \right), \\
        \overset{BCH}{\to} \rm{Exp} \left( \mathbf{E} \right) \ \rm{Exp} \left( \mathcal{J}_l\left(\mathbf{E}\right)^{T} \delta \mathbf{E} \right) 
        &= \mathbf{d}^T \mathbf{a} \ \rm{Exp} \left(\mathbf{b} \right) \ \rm{Exp} \left( \mathcal{J}_l\left(\mathbf{b}\right)^{T} \delta \mathbf{b} \right) \ \rm{Exp} \left( \mathbf{c} \right), \\
        \to \rm{Exp} \left( \mathcal{J}_l\left(\mathbf{E}\right)^{T} \delta \mathbf{E} \right) 
        &= \rm{Exp} \left( \mathbf{c} \right)^T \ \rm{Exp} \left( \mathcal{J}_l\left(\mathbf{b}\right)^{T} \delta \mathbf{b} \right) \ \rm{Exp} \left( \mathbf{c} \right) \\
        &\overset{\textbf{Adjoint 1}}{=} \rm{Exp} \left( \rm{Exp} \left( \mathbf{c} \right)^T \ \mathcal{J}_l\left(\mathbf{b}\right)^{T} \delta \mathbf{b} \right), \\
        \to  \mathcal{J}_l\left(\mathbf{E}\right)^{T} \delta \mathbf{E} 
        &= \rm{Exp} \left( -\mathbf{c} \right) \ \mathcal{J}_l\left(\mathbf{b}\right)^{T} \delta \mathbf{b}, \\
        \to \frac{\partial \mathbf{E}}{\partial \mathbf{b}} 
        &= \mathcal{J}_l\left(\mathbf{E}\right)^{-T} \rm{Exp} \left( -\mathbf{c} \right) \ \mathcal{J}_l\left(\mathbf{b}\right)^{T}
    \end{aligned}
\end{equation}
where $\mathcal{J}_l \left( \mathbf{u} \right) = \frac{\sin \theta}{\theta} \mathbf{I} + \left( 1 - \frac{\sin \theta}{\theta}\right) \mathbf{a}\mathbf{a}^T + \frac{1- \cos \theta}{\theta} \lfloor\mathbf{a}\rfloor_{{\wedge}}, \mathbf{u} = \theta \mathbf{a}, \|\mathbf{a}\|=1$ and $\textbf{Adjoint 1}$ represents $\mathbf{R} \rm{Exp} \left(\mathbf{p}\right) \mathbf{R}^T = \rm{Exp} \left(\mathbf{R}\mathbf{p}\right), \ \mathbf{R} \in \rm{SO}\left(3\right), \mathbf{p} \in \mathbb{R}_3 $.
$\frac{\partial \mathbf{E}}{\partial \mathbf{c}}$ is similarly derived to 
\begin{equation}
\label{e:manifold_rules_so3_c}
    \begin{aligned}
       \rm{Exp} \left( \mathbf{E} + \delta \mathbf{E} \right) &= \mathbf{d}^T \mathbf{a} \ \rm{Exp} \left(\mathbf{b}\right) \ \rm{Exp} \left( \mathbf{c} + \delta \mathbf{c} \right), \\
       \overset{BCH}{\to} \rm{Exp} \left( \mathbf{E} \right) \ \rm{Exp} \left( \mathcal{J}_l\left(\mathbf{E}\right)^{T} \delta \mathbf{E} \right) 
       &= \mathbf{d}^T \mathbf{a} \ \rm{Exp} \left(\mathbf{b} \right) \ \rm{Exp} \left( \mathbf{c} \right) \ \rm{Exp} \left( \mathcal{J}_l\left(\mathbf{c}\right)^{T} \delta \mathbf{c} \right),\\
       \to \rm{Exp} \left( \mathcal{J}_l\left(\mathbf{E}\right)^{T} \delta \mathbf{E} \right) 
       &=  \rm{Exp} \left( \mathcal{J}_l\left(\mathbf{c}\right)^{T} \delta \mathbf{c} \right),\\
       \to \mathcal{J}_l\left(\mathbf{E}\right)^{T} \delta \mathbf{E} 
       &= \mathcal{J}_l\left(\mathbf{c}\right)^{T} \delta \mathbf{c}, \\
       \to \frac{\partial \mathbf{E}}{\partial \mathbf{c}} 
       &= \mathcal{J}_l\left(\mathbf{E}\right)^{-T} \mathcal{J}_l\left(\mathbf{c}\right)^{T}.
    \end{aligned}
\end{equation}
Referring to (\ref{e:manifold_rules}) and (\ref{e:manifold_rules_so3_b}), 
\begin{equation}
\label{e:fx_part1}
    \begin{aligned}
        \left.\frac{\partial  \mathbf{G}\left(\widetilde{\mathbf{x}}_i, \mathbf{g}\left(\mathbf{0}, \mathbf{0}\right)\right)}{\partial \widetilde{\mathbf{x}}_i}\right|_{\widetilde{\mathbf{x}}_i = \mathbf{0}} 
        &= 
        \begin{bmatrix}
            \mathcal{J}_l\left(\mathbf{0}\right)^{-1} \rm{Exp} \left( -\mathbf{g} \left(\mathbf{0,0}\right) \right) \ \mathcal{J}_l\left(\mathbf{0}\right)^{T} & \mathbf{0} \\
            \mathbf{0} & \mathbf{I}_{15\times15}
        \end{bmatrix} \\
        &= \begin{bmatrix}
            \rm{Exp} \left( -\Delta t \mathbf{f}\left( \widehat{\mathbf{x}}_i, \mathbf{u}_i, \mathbf{0} \right) \right) & \mathbf{0} \\
            \mathbf{0} & \mathbf{I}_{15\times15}
        \end{bmatrix} \\
        &= \begin{bmatrix}
            \rm{Exp} \left( -\Delta t \left( \bm{\omega}_{m_i} -  \widehat{\mathbf{b}}_{\bm{\omega}_i} \right) \right) & \mathbf{0} \\
            \mathbf{0} & \mathbf{I}_{15\times15}
        \end{bmatrix}.
    \end{aligned}
\end{equation}
Referring to (\ref{e:manifold_rules}) and (\ref{e:manifold_rules_so3_c}),
\begin{equation}
\label{e:fx_part2}
    \begin{aligned}
        \left.\frac{\partial  \mathbf{G}\left(\mathbf{0}, \mathbf{g}\left(\widetilde{\mathbf{x}}_i, \mathbf{0}\right)\right)}{\partial \mathbf{g}\left(\widetilde{\mathbf{x}}_i, \mathbf{0}\right)}\right|_{\widetilde{\mathbf{x}}_i = \mathbf{0}} 
        &= 
        \begin{bmatrix}
            \mathcal{J}_l\left(\mathbf{0}\right)^{-1}\mathcal{J}_l\left(\mathbf{g} \left(\mathbf{0,0}\right)\right)^{T} & \mathbf{0} \\
            \mathbf{0} & \mathbf{I}_{15\times15}
        \end{bmatrix}  \\
        &= 
        \begin{bmatrix}
            \mathcal{J}_l\left(\Delta t \left( \bm{\omega}_{m_i} -  \widehat{\mathbf{b}}_{\bm{\omega}_i} \right)\right)^{T} & \mathbf{0} \\
            \mathbf{0} & \mathbf{I}_{15\times15}
        \end{bmatrix}. 
    \end{aligned}
\end{equation}
As $\mathbf{g}\left(\widetilde{\mathbf{x}}_i, \mathbf{w}_i\right) = \Delta t \mathbf{f}\left(\widehat{\mathbf{x}}_i \boxplus \widetilde{\mathbf{x}}_i, \mathbf{u}_i, \mathbf{w}_i \right)$, thus
\begin{equation}
\label{e:fx_part3}
    \begin{aligned}
        \mathbf{g}\left(\widetilde{\mathbf{x}}_i, \mathbf{w}_i\right) 
        &= \Delta t \begin{bmatrix} \bm{\omega}_{m_i} - \mathbf{b}_{\bm{\omega}_i} - \mathbf{n}_{\bm{\omega}_i} \\
        {}^G_{I_i} \mathbf{v} \\
        {}^G_{I_i} \mathbf{R} \left(\mathbf{a}_{m_i} - \mathbf{b}_{\mathbf{a}_i} - \mathbf{n}_{\mathbf{a}_i}\right) + {}^G \mathbf{g}_i  \\
        \mathbf{n}_{\mathbf{b} \bm{\omega}_i} \\
        \mathbf{n}_{\mathbf{b} \mathbf{a}_i} \\
        \mathbf{0}_{3\times1} \end{bmatrix} 
        = \Delta t \begin{bmatrix} \bm{\omega}_{m_i} - \widehat{\mathbf{b}}_{\bm{\omega}_i} - \widetilde{\mathbf{b}}_{\bm{\omega}_i} - \mathbf{n}_{\bm{\omega}_i} \\
        {}^G_{I_i} \widehat{\mathbf{v}} +  {}^G_{I_i} \widetilde{\mathbf{v}}\\
        {}^G_{I_i} \widehat{\mathbf{R}} \ \rm{Exp} \left( \delta \bm{\theta}_i \right) \left(\mathbf{a}_{m_i} - \widehat{\mathbf{b}}_{\mathbf{a}_i} - \widetilde{\mathbf{b}}_{\mathbf{a}_i} - \mathbf{n}_{\mathbf{a}_i}\right) + {}^G \mathbf{g}_i  \\
        \mathbf{n}_{\mathbf{b} \bm{\omega}_i} \\
        \mathbf{n}_{\mathbf{b} \mathbf{a}_i} \\
        \mathbf{0}_{3\times1} \end{bmatrix}, \\
        \to \mathbf{g}\left(\widetilde{\mathbf{x}}_i, \mathbf{0}\right) 
        &= \Delta t \begin{bmatrix} \bm{\omega}_{m_i} - \widehat{\mathbf{b}}_{\bm{\omega}_i} - \widetilde{\mathbf{b}}_{\bm{\omega}_i}  \\
        {}^G_{I_i} \widehat{\mathbf{v}} +  {}^G_{I_i} \widetilde{\mathbf{v}}\\
        {}^G_{I_i} \widehat{\mathbf{R}} \ \rm{Exp} \left( \delta \bm{\theta}_i \right) \left(\mathbf{a}_{m_i} - \widehat{\mathbf{b}}_{\mathbf{a}_i} - \widetilde{\mathbf{b}}_{\mathbf{a}_i}\right) + {}^G \mathbf{g}_i  \\
        \mathbf{0}_{3\times1} \\
        \mathbf{0}_{3\times1} \\
        \mathbf{0}_{3\times1} \end{bmatrix} \\
        &= \Delta t \begin{bmatrix} \bm{\omega}_{m_i} - \widehat{\mathbf{b}}_{\bm{\omega}_i} - \widetilde{\mathbf{b}}_{\bm{\omega}_i}  \\
        {}^G_{I_i} \widehat{\mathbf{v}} +  {}^G_{I_i} \widetilde{\mathbf{v}}\\
        {}^G_{I_i} \widehat{\mathbf{R}} \left(\mathbf{a}_{m_i} - \widehat{\mathbf{b}}_{\mathbf{a}_i} - \widetilde{\mathbf{b}}_{\mathbf{a}_i}\right) - {}^G_{I_i} \widehat{\mathbf{R}} \lfloor\mathbf{a}_{m_i} - \widehat{\mathbf{b}}_{\mathbf{a}_i} - \widetilde{\mathbf{b}}_{\mathbf{a}_i}\rfloor_{\wedge} \delta \bm{\theta}_i + {}^G \mathbf{g}_i  \\
        \mathbf{0}_{3\times1} \\
        \mathbf{0}_{3\times1} \\
        \mathbf{0}_{3\times1} \end{bmatrix}, \\
        \to \left.\frac{\partial  \mathbf{g}\left(\widetilde{\mathbf{x}}_i, \mathbf{0}\right) }{\partial \widetilde{\mathbf{x}}_i }\right|_{ \widetilde{\mathbf{x}}_i = \mathbf{0}} 
        &= \begin{bmatrix}
            \mathbf{0} &   \mathbf{0} &  \mathbf{0} & -\mathbf{I}_{3\times3} \Delta t &  \mathbf{0} &  \mathbf{0} \\
            \mathbf{0} &   \mathbf{0} &  \mathbf{I}_{3\times3} \Delta t & \mathbf{0}  &  \mathbf{0} &  \mathbf{0} \\
            - {}^G_{I_i} \widehat{\mathbf{R}} \lfloor\mathbf{a}_{m_i} - \widehat{\mathbf{b}}_{\mathbf{a}_i} \rfloor_{\wedge} \Delta t & \mathbf{0} &  \mathbf{0} & \mathbf{0}  & - {}^G_{I_i} \widehat{\mathbf{R}} \Delta t & \mathbf{I}_{3\times3} \Delta t\\ 
            \mathbf{0} &   \mathbf{0} &  \mathbf{0} & \mathbf{0}  &  \mathbf{0} &  \mathbf{0} \\
            \mathbf{0} &   \mathbf{0} &  \mathbf{0} & \mathbf{0}  &  \mathbf{0} &  \mathbf{0} \\
            \mathbf{0} &   \mathbf{0} &  \mathbf{0} & \mathbf{0}  &  \mathbf{0} &  \mathbf{0} \\
        \end{bmatrix}.
    \end{aligned}
\end{equation}
Referring to (\ref{e:fx_part1}), (\ref{e:fx_part2}) and (\ref{e:fx_part3}), we can derive the exact formulation of $\mathbf{F}_{\widetilde{\mathbf{x}}} $
\begin{equation}
\label{e:F_partial_x_exact}
    \begin{aligned}
        \mathbf{F}_{\widetilde{\mathbf{x}}} 
        &= \left.\frac{\partial  \mathbf{G}\left(\widetilde{\mathbf{x}}_i, \mathbf{g}\left(\mathbf{0}, \mathbf{0}\right)\right)}{\partial \widetilde{\mathbf{x}}_i} 
        + \frac{\partial  \mathbf{G}\left(\mathbf{0}, \mathbf{g}\left(\widetilde{\mathbf{x}}_i, \mathbf{0}\right)\right)}{\partial \mathbf{g}\left(\widetilde{\mathbf{x}}_i, \mathbf{0}\right)} \frac{\partial \mathbf{g}\left(\widetilde{\mathbf{x}}_i, \mathbf{0}\right)}{\partial \widetilde{\mathbf{x}}_i} \right|_{\widetilde{\mathbf{x}}_i = \mathbf{0}} \\
        &= \begin{bmatrix}
            \rm{Exp} \left( -\Delta t \left( \bm{\omega}_{m_i} -  \widehat{\mathbf{b}}_{\bm{\omega}_i} \right) \right) & \mathbf{0} \\
            \mathbf{0} & \mathbf{I}_{15\times15}
        \end{bmatrix} \\
        &+ \begin{bmatrix}
            \mathcal{J}_l\left(\Delta t \left( \bm{\omega}_{m_i} -  \widehat{\mathbf{b}}_{\bm{\omega}_i} \right)\right)^{T} & \mathbf{0} \\
            \mathbf{0} & \mathbf{I}_{15\times15}
        \end{bmatrix} \begin{bmatrix}
            \mathbf{0} &   \mathbf{0} &  \mathbf{0} & -\mathbf{I}_{3\times3} \Delta t &  \mathbf{0} &  \mathbf{0} \\
            \mathbf{0} &   \mathbf{0} &  \mathbf{I}_{3\times3} \Delta t & \mathbf{0}  &  \mathbf{0} &  \mathbf{0} \\
            - {}^G_{I_i} \widehat{\mathbf{R}} \lfloor\mathbf{a}_{m_i} - \widehat{\mathbf{b}}_{\mathbf{a}_i} \rfloor_{\wedge} \Delta t & \mathbf{0} &  \mathbf{0} & \mathbf{0}  & - {}^G_{I_i} \widehat{\mathbf{R}} \Delta t & \mathbf{I}_{3\times3} \Delta t\\ 
            \mathbf{0} &   \mathbf{0} &  \mathbf{0} & \mathbf{0}  &  \mathbf{0} &  \mathbf{0} \\
            \mathbf{0} &   \mathbf{0} &  \mathbf{0} & \mathbf{0}  &  \mathbf{0} &  \mathbf{0} \\
            \mathbf{0} &   \mathbf{0} &  \mathbf{0} & \mathbf{0}  &  \mathbf{0} &  \mathbf{0} \\
        \end{bmatrix} \\
        &= \begin{bmatrix}
            \rm{Exp} \left( -\Delta t \left( \bm{\omega}_{m_i} -  \widehat{\mathbf{b}}_{\bm{\omega}_i} \right) \right) &   \mathbf{0} &  \mathbf{0} & -\mathcal{J}_l\left(\Delta t \left( \bm{\omega}_{m_i} -  \widehat{\mathbf{b}}_{\bm{\omega}_i} \right)\right)^{T} \Delta t &  \mathbf{0} &  \mathbf{0} \\
            \mathbf{0} &   \mathbf{I}_{3\times3} &  \mathbf{I}_{3\times3} \Delta t & \mathbf{0}  &  \mathbf{0} &  \mathbf{0} \\
            - {}^G_{I_i} \widehat{\mathbf{R}} \lfloor\mathbf{a}_{m_i} - \widehat{\mathbf{b}}_{\mathbf{a}_i} \rfloor_{\wedge} \Delta t & \mathbf{0} &  \mathbf{I}_{3\times3} & \mathbf{0}  & - {}^G_{I_i} \widehat{\mathbf{R}} \Delta t & \mathbf{I}_{3\times3} \Delta t\\ 
            \mathbf{0} &   \mathbf{0} &  \mathbf{0} & \mathbf{I}_{3\times3}  &  \mathbf{0} &  \mathbf{0} \\
            \mathbf{0} &   \mathbf{0} &  \mathbf{0} & \mathbf{0}  &  \mathbf{I}_{3\times3} &  \mathbf{0} \\
            \mathbf{0} &   \mathbf{0} &  \mathbf{0} & \mathbf{0}  &  \mathbf{0} &  \mathbf{I}_{3\times3} \\
        \end{bmatrix} \\
        & \approx \begin{bmatrix}
            \rm{Exp} \left( -\Delta t \left( \bm{\omega}_{m_i} -  \widehat{\mathbf{b}}_{\bm{\omega}_i} \right) \right) &   \mathbf{0} &  \mathbf{0} & -\mathbf{I}_{3\times3} \Delta t &  \mathbf{0} &  \mathbf{0} \\
            \mathbf{0} &   \mathbf{I}_{3\times3} &  \mathbf{I}_{3\times3} \Delta t & \mathbf{0}  &  \mathbf{0} &  \mathbf{0} \\
            - {}^G_{I_i} \widehat{\mathbf{R}} \lfloor\mathbf{a}_{m_i} - \widehat{\mathbf{b}}_{\mathbf{a}_i} \rfloor_{\wedge} \Delta t & \mathbf{0} &  \mathbf{I}_{3\times3} & \mathbf{0}  & - {}^G_{I_i} \widehat{\mathbf{R}} \Delta t & \mathbf{I}_{3\times3} \Delta t\\ 
            \mathbf{0} &   \mathbf{0} &  \mathbf{0} & \mathbf{I}_{3\times3}  &  \mathbf{0} &  \mathbf{0} \\
            \mathbf{0} &   \mathbf{0} &  \mathbf{0} & \mathbf{0}  &  \mathbf{I}_{3\times3} &  \mathbf{0} \\
            \mathbf{0} &   \mathbf{0} &  \mathbf{0} & \mathbf{0}  &  \mathbf{0} &  \mathbf{I}_{3\times3} \\
        \end{bmatrix}.
    \end{aligned}
\end{equation}
Similarly referring to (\ref{e:manifold_rules}) and (\ref{e:manifold_rules_so3_c}),
\begin{equation}
\label{e:fw_part1}
    \begin{aligned}
        \left.\frac{\partial  \mathbf{G}\left(\mathbf{0}, \mathbf{g}\left(\mathbf{0}, \mathbf{w}_i\right)\right)}{\partial \mathbf{g}\left(\mathbf{0}, \mathbf{w}_i\right)}\right|_{\mathbf{w}_i =\mathbf{0}} 
        &= 
        \begin{bmatrix}
            \mathcal{J}_l\left(\mathbf{0}\right)^{-1}\mathcal{J}_l\left(\mathbf{g} \left(\mathbf{0,0}\right)\right)^{T} & \mathbf{0} \\
            \mathbf{0} & \mathbf{I}_{15\times15}
        \end{bmatrix}  \\
        &= 
        \begin{bmatrix}
            \mathcal{J}_l\left(\Delta t \left( \bm{\omega}_{m_i} -  \widehat{\mathbf{b}}_{\bm{\omega}_i} \right)\right)^{T} & \mathbf{0} \\
            \mathbf{0} & \mathbf{I}_{15\times15}
        \end{bmatrix} . 
    \end{aligned}
\end{equation}
Like (\ref{e:fx_part3}), 
\begin{equation}
\label{e:fw_part2}
    \begin{aligned}
        \mathbf{g}\left(\mathbf{0}, \mathbf{w}_i\right)
        &= 
        \Delta t \begin{bmatrix} \bm{\omega}_{m_i} - \widehat{\mathbf{b}}_{\bm{\omega}_i}  - \mathbf{n}_{\bm{\omega}_i} \\
        {}^G_{I_i} \widehat{\mathbf{v}}\\
        {}^G_{I_i} \widehat{\mathbf{R}} \left(\mathbf{a}_{m_i} - \widehat{\mathbf{b}}_{\mathbf{a}_i} - \mathbf{n}_{\mathbf{a}_i}\right) + {}^G \widehat{\mathbf{g}}_i  \\
        \mathbf{n}_{\mathbf{b} \bm{\omega}_i} \\
        \mathbf{n}_{\mathbf{b} \mathbf{a}_i} \\
        \mathbf{0}_{3\times1} \end{bmatrix},
        \\
        \left.\frac{\partial \mathbf{g}\left(\mathbf{0}, \mathbf{w}_i\right)}{\partial \mathbf{w}_i} \right|_{\mathbf{w}_i =\mathbf{0}} 
        &=
        \begin{bmatrix}
            -\mathbf{I}_{3\times3} \Delta t & \mathbf{0} & \mathbf{0} & \mathbf{0} \\
            \mathbf{0} & \mathbf{0} & \mathbf{0} & \mathbf{0} \\
            \mathbf{0} & -{}^G_{I_i} \widehat{\mathbf{R}} \Delta t & \mathbf{0} & \mathbf{0} \\
            \mathbf{0} & \mathbf{0} & \mathbf{I}_{3\times3} \Delta t & \mathbf{0} \\
            \mathbf{0} & \mathbf{0} & \mathbf{0} & \mathbf{I}_{3\times3} \Delta t \\
            \mathbf{0} & \mathbf{0} & \mathbf{0} & \mathbf{0}
        \end{bmatrix}.
    \end{aligned}
\end{equation}
Combine (\ref{e:fw_part1}) with (\ref{e:fw_part2}), derive $\mathbf{F}_{\mathbf{w}} $
\begin{equation}
\label{e:F_partial_w_exact}
    \begin{aligned}
        \mathbf{F}_{\mathbf{w}}
        &= \begin{bmatrix}
            \mathcal{J}_l\left(\Delta t \left( \bm{\omega}_{m_i} -  \widehat{\mathbf{b}}_{\bm{\omega}_i} \right)\right)^{T} & \mathbf{0} \\
            \mathbf{0} & \mathbf{I}_{15\times15}
        \end{bmatrix} \begin{bmatrix}
            -\mathbf{I}_{3\times3} \Delta t & \mathbf{0} & \mathbf{0} & \mathbf{0} \\
            \mathbf{0} & \mathbf{0} & \mathbf{0} & \mathbf{0} \\
            \mathbf{0} & -{}^G_{I_i} \widehat{\mathbf{R}} \Delta t & \mathbf{0} & \mathbf{0} \\
            \mathbf{0} & \mathbf{0} & \mathbf{I}_{3\times3} \Delta t & \mathbf{0} \\
            \mathbf{0} & \mathbf{0} & \mathbf{0} & \mathbf{I}_{3\times3} \Delta t \\
            \mathbf{0} & \mathbf{0} & \mathbf{0} & \mathbf{0}
        \end{bmatrix} \\
        &= \begin{bmatrix}
            -\mathcal{J}_l\left(\Delta t \left( \bm{\omega}_{m_i} -  \widehat{\mathbf{b}}_{\bm{\omega}_i} \right)\right)^{T}\Delta t & \mathbf{0} & \mathbf{0} & \mathbf{0} \\
            \mathbf{0} & \mathbf{0} & \mathbf{0} & \mathbf{0} \\
            \mathbf{0} & -{}^G_{I_i} \widehat{\mathbf{R}} \Delta t & \mathbf{0} & \mathbf{0} \\
            \mathbf{0} & \mathbf{0} & \mathbf{I}_{3\times3} \Delta t & \mathbf{0} \\
            \mathbf{0} & \mathbf{0} & \mathbf{0} & \mathbf{I}_{3\times3} \Delta t \\
            \mathbf{0} & \mathbf{0} & \mathbf{0} & \mathbf{0}
        \end{bmatrix} \\
        &\approx \begin{bmatrix}
            -\mathbf{I}_{3\times3}\Delta t & \mathbf{0} & \mathbf{0} & \mathbf{0} \\
            \mathbf{0} & \mathbf{0} & \mathbf{0} & \mathbf{0} \\
            \mathbf{0} & -{}^G_{I_i} \widehat{\mathbf{R}} \Delta t & \mathbf{0} & \mathbf{0} \\
            \mathbf{0} & \mathbf{0} & \mathbf{I}_{3\times3} \Delta t & \mathbf{0} \\
            \mathbf{0} & \mathbf{0} & \mathbf{0} & \mathbf{I}_{3\times3} \Delta t \\
            \mathbf{0} & \mathbf{0} & \mathbf{0} & \mathbf{0}
        \end{bmatrix}.
    \end{aligned}
\end{equation}

\subsubsection{Method 2: Continuous-time Error Dynamics and Discretization}
According to (\ref{e:kine_model_continuous}) and (\ref{e:error_state_def}), we can easily derive the dynamics of $\delta \mathbf{t}, \delta \mathbf{b}_{\bm{\omega}}, \delta \mathbf{b}_{\mathbf{a}}$ and $\delta {}^G \mathbf{g}$
\begin{equation}
\label{e:ct_err_dynamics_part1}
    \begin{aligned}
        \delta {}^G_I \dot{\mathbf{t}} = \delta {}^G_I \mathbf{v},
        \ \delta {}^G \dot{\mathbf{g}} = \mathbf{0},
        \ \delta \dot{\mathbf{b}}_{\bm{\omega}} = \mathbf{n}_{\mathbf{b}\bm{\omega}}, 
        \ \delta \dot{\mathbf{b}}_{\mathbf{a}} = \mathbf{n}_{\mathbf{b} \mathbf{a}}.
    \end{aligned}
\end{equation}
For the dynamic of $\delta \bm{\theta}$, 
\begin{equation}
\label{e:ct_err_dynamics_pose}
    \begin{aligned}
        {}^G_I \dot{\mathbf{R}} 
        &= {}^G_I \mathbf{R} \lfloor \bm{\omega}_m -  \mathbf{b}_{\bm{\omega}} - \mathbf{n}_{\bm{\omega}} \rfloor_{\wedge}, \\
        {}^G_I \dot{\widehat{\mathbf{R}}} 
        &= {}^G_I \widehat{\mathbf{R}} \lfloor \bm{\omega}_m -  \widehat{\mathbf{b}}_{\bm{\omega}} \rfloor_{\wedge}, \\
        {}^G_I \mathbf{R} 
        &= {}^G_I \widehat{\mathbf{R}} \rm{Exp} \left(\delta \bm{\theta} \right),  \\
        \to{}^G_I \dot{\mathbf{R}} 
        &= {}^G_I \dot{\widehat{\mathbf{R}}} \ \rm{Exp} \left(\delta \bm{\theta} \right) + {}^G_I \widehat{\mathbf{R}} \ \rm{Exp} \left(\delta \bm{\theta} \right)  \lfloor \delta \dot{\bm{\theta}} \rfloor_{\wedge}, \\
        \to {}^G_I \widehat{\mathbf{R}} \ \rm{Exp} \left(\delta \bm{\theta} \right)\lfloor \bm{\omega}_m -  \mathbf{b}_{\bm{\omega}} - \mathbf{n}_{\bm{\omega}} \rfloor_{\wedge} 
        &=  {}^G_I \widehat{\mathbf{R}} \lfloor \bm{\omega}_m -  \widehat{\mathbf{b}}_{\bm{\omega}} \rfloor_{\wedge} \ \rm{Exp} \left(\delta \bm{\theta} \right) + {}^G_I \widehat{\mathbf{R}} \ \rm{Exp} \left(\delta \bm{\theta} \right)  \lfloor \delta \dot{\bm{\theta}} \rfloor_{\wedge}, \\
        \to \rm{Exp} \left(\delta \bm{\theta} \right)\lfloor \bm{\omega}_m -  \mathbf{b}_{\bm{\omega}} - \mathbf{n}_{\bm{\omega}} \rfloor_{\wedge} 
        &= \lfloor \bm{\omega}_m -  \widehat{\mathbf{b}}_{\bm{\omega}} \rfloor_{\wedge} \ \rm{Exp} \left(\delta \bm{\theta} \right) + \rm{Exp} \left(\delta \bm{\theta} \right)  \lfloor \delta \dot{\bm{\theta}} \rfloor_{\wedge}, \\
        \to \rm{Exp} \left(\delta \bm{\theta} \right)  \lfloor \delta \dot{\bm{\theta}} \rfloor_{\wedge} 
        &= \rm{Exp} \left(\delta \bm{\theta} \right)\lfloor \bm{\omega}_m -  \mathbf{b}_{\bm{\omega}} - \mathbf{n}_{\bm{\omega}} \rfloor_{\wedge} - \lfloor \bm{\omega}_m -  \widehat{\mathbf{b}}_{\bm{\omega}} \rfloor_{\wedge} \ \rm{Exp} \left(\delta \bm{\theta} \right) \\
        &\overset{\textbf{Adjoint 2}}{=} \rm{Exp} \left(\delta \bm{\theta} \right)\lfloor \bm{\omega}_m -  \mathbf{b}_{\bm{\omega}} - \mathbf{n}_{\bm{\omega}} \rfloor_{\wedge} - \rm{Exp} \left(\delta \bm{\theta} \right) \lfloor \rm{Exp} \left(-\delta \bm{\theta} \right) \left( \bm{\omega}_m -  \widehat{\mathbf{b}}_{\bm{\omega}} \right) \rfloor_{\wedge}, \\
        \to \delta \dot{\bm{\theta}} 
        &= \left( \bm{\omega}_m -  \mathbf{b}_{\bm{\omega}} - \mathbf{n}_{\bm{\omega}} \right) - \rm{Exp} \left(-\delta \bm{\theta} \right) \left( \bm{\omega}_m -  \widehat{\mathbf{b}}_{\bm{\omega}} \right) \\
        &\approx \bm{\omega}_m -  \widehat{\mathbf{b}}_{\bm{\omega}} -  \widetilde{\mathbf{b}}_{\bm{\omega}} - \mathbf{n}_{\bm{\omega}} - \left(\mathbf{I}- \lfloor \delta \bm{\theta} \rfloor_{\wedge} \right)\left( \bm{\omega}_m -  \widehat{\mathbf{b}}_{\bm{\omega}} \right) \\
        &= - \lfloor \bm{\omega}_m -  \widehat{\mathbf{b}}_{\bm{\omega}} \rfloor_{\wedge} \delta \bm{\theta} -  \widetilde{\mathbf{b}}_{\bm{\omega}} - \mathbf{n}_{\bm{\omega}}
    \end{aligned}
\end{equation}
where $\textbf{Adjoint 2}$ denotes $ \lfloor\mathbf{p}\rfloor_{\wedge} \mathbf{R} = \mathbf{R} \lfloor \mathbf{R}^T \mathbf{p} \rfloor_{\wedge}, \ \mathbf{R} \in \rm{SO}\left(3\right), \mathbf{p} \in \mathbb{R}_3$.
For the dynamic of $\delta {}^G_I \dot{\mathbf{v}}$,
\begin{equation}
\label{e:ct_err_dynamics_vel}
    \begin{aligned}
        {}^G_I \dot{\mathbf{v}} 
        &= {}^G_I \mathbf{R} \left( \mathbf{a}_m - \mathbf{b}_{\mathbf{a}} - \mathbf{n}_{\mathbf{a}} \right) + {}^G \mathbf{g}, \\
        {}^G_I \dot{\widehat{\mathbf{v}}} 
        &= {}^G_I \widehat{\mathbf{R}} \left( \mathbf{a}_m - \widehat{\mathbf{b}}_{\mathbf{a}} \right) + {}^G \widehat{\mathbf{g}}, \\
        {}^G_I \mathbf{v} 
        &= {}^G_I \widehat{\mathbf{v}} + {}^G_I \widetilde{\mathbf{v}}, \\
        \to {}^G_I \dot{\mathbf{v}} 
        &= {}^G_I \dot{\widehat{\mathbf{v}}} + {}^G_I \dot{\widetilde{\mathbf{v}}}, \\
        \to {}^G_I \mathbf{R} \left( \mathbf{a}_m - \mathbf{b}_{\mathbf{a}} - \mathbf{n}_{\mathbf{a}} \right) + {}^G \mathbf{g} 
        &={}^G_I \widehat{\mathbf{R}} \left( \mathbf{a}_m - \widehat{\mathbf{b}}_{\mathbf{a}} \right) + {}^G \widehat{\mathbf{g}} + {}^G_I \dot{\widetilde{\mathbf{v}}}, \\
        {}^G_I \dot{\widetilde{\mathbf{v}}} 
        &= {}^G_I \widehat{\mathbf{R}} \left(\mathbf{I} + \lfloor\delta\bm{\theta}\rfloor_{\wedge}\right) \left( \mathbf{a}_m - \mathbf{b}_{\mathbf{a}} - \mathbf{n}_{\mathbf{a}} \right) + {}^G \mathbf{g} - {}^G_I \widehat{\mathbf{R}} \left( \mathbf{a}_m - \widehat{\mathbf{b}}_{\mathbf{a}} \right) - {}^G \widehat{\mathbf{g}} \\
        &= {}^G_I \widehat{\mathbf{R}} \left(-\lfloor \mathbf{a}_m - \mathbf{b}_{\mathbf{a}} - \mathbf{n}_{\mathbf{a}} \rfloor_{\wedge} \delta \bm{\theta} - \widetilde{\mathbf{b}}_{\mathbf{a}} - \mathbf{n}_{\mathbf{a}} \right) + {}^G \widetilde{\mathbf{g}} \\
        &\approx -{}^G_I \widehat{\mathbf{R}} \lfloor \mathbf{a}_m - \widehat{\mathbf{b}}_{\mathbf{a}} \rfloor_{\wedge} \delta \bm{\theta} - {}^G_I \widehat{\mathbf{R}} \widetilde{\mathbf{b}}_{\mathbf{a}} - {}^G_I \widehat{\mathbf{R}}\mathbf{n}_{\mathbf{a}} + {}^G \widetilde{\mathbf{g}}.
    \end{aligned}
\end{equation}
Take the discretization for continuous-time error dynamics,
\begin{equation}
\label{e:dt_err_dynamics}
    \begin{aligned}
        \delta \bm{\theta}_{i+1} &= \rm{Exp}\left(- \left( \bm{\omega}_{m_i} -  \widehat{\mathbf{b}}_{\bm{\omega}_i} \right) \Delta t \right) \delta \bm{\theta}_i -  \widetilde{\mathbf{b}}_{\bm{\omega}_i}  \Delta t  - \mathbf{n}_{\bm{\omega}_i}  \Delta t, \\
        {}^G_{I_{i+1}} \widetilde{\mathbf{t}} &= {}^G_{I_i} \widetilde{\mathbf{t}} + {}^G_{I_i} \widetilde{\mathbf{v}} \Delta t, \\
        {}^G_{I_{i+1}} \widetilde{\mathbf{v}} &= {}^G_{I_i} \widetilde{\mathbf{v}} + \left( -{}^G_{I_i} \widehat{\mathbf{R}} \lfloor \mathbf{a}_{m_i} - \widehat{\mathbf{b}}_{\mathbf{a}_i} \rfloor_{\wedge} \delta \bm{\theta}_i - {}^G_{I_i} \widehat{\mathbf{R}} \widetilde{\mathbf{b}}_{\mathbf{a}_i} - {}^G_{I_i} \widehat{\mathbf{R}}\mathbf{n}_{\mathbf{a}_i} + {}^G \widetilde{\mathbf{g}}_i \right) \Delta t, \\
        \widetilde{\mathbf{b}}_{\bm{\omega}_{i+1}} &= \widetilde{\mathbf{b}}_{\bm{\omega}_i} + \mathbf{n}_{\mathbf{b}\bm{\omega}_i} \Delta t, \\
        \widetilde{\mathbf{b}}_{\mathbf{a}_{i+1}} &= \widetilde{\mathbf{b}}_{\mathbf{a}_i} + \mathbf{n}_{\mathbf{b} \mathbf{a}_i} \Delta t, \\
        {}^G \widetilde{\mathbf{g}}_{i+1} &= {}^G \widetilde{\mathbf{g}}_i.
    \end{aligned}
\end{equation}
Recall the error state model in (\ref{e:error_state_model}),
\begin{equation}
\label{e:ct_fx_fw}
    \begin{aligned}
        \mathbf{F}_{\widetilde{\mathbf{x}}} 
        &= 
        \begin{bmatrix}
            \rm{Exp}\left(- \left( \bm{\omega}_{m_i} -  \widehat{\mathbf{b}}_{\bm{\omega}_i} \right) \Delta t \right) & \mathbf{0} &  \mathbf{0} & -\mathbf{I}_{3\times3} \Delta t & \mathbf{0} &  \mathbf{0} \\
            \mathbf{0} & \mathbf{I}_{3\times3} & \mathbf{I}_{3\times3} \Delta t &   \mathbf{0} & \mathbf{0} &  \mathbf{0} \\
            -{}^G_{I_i} \widehat{\mathbf{R}} \lfloor \mathbf{a}_{m_i} - \widehat{\mathbf{b}}_{\mathbf{a}_i} \rfloor_{\wedge} \Delta t & \mathbf{0} & \mathbf{I}_{3\times3} & \mathbf{0} & -{}^G_{I_i} \widehat{\mathbf{R}} \Delta t & \mathbf{I}_{3\times3} \Delta t \\
             \mathbf{0} &   \mathbf{0} &  \mathbf{0} & \mathbf{I}_{3\times3}  &  \mathbf{0} &  \mathbf{0} \\
            \mathbf{0} &   \mathbf{0} &  \mathbf{0} & \mathbf{0}  &  \mathbf{I}_{3\times3} &  \mathbf{0} \\
            \mathbf{0} &   \mathbf{0} &  \mathbf{0} & \mathbf{0}  &  \mathbf{0} &  \mathbf{I}_{3\times3} \\
        \end{bmatrix}, \\
        \mathbf{F}_{\mathbf{w}} 
        &= 
        \begin{bmatrix}
            -\mathbf{I}_{3\times3} \Delta t & \mathbf{0} &   \mathbf{0} &  \mathbf{0} \\
            \mathbf{0} &   \mathbf{0} &  \mathbf{0} & \mathbf{0} \\
            \mathbf{0} & -{}^G_{I_i} \widehat{\mathbf{R}} \Delta t &  \mathbf{0} & \mathbf{0} \\
            \mathbf{0} &   \mathbf{0} &  \mathbf{I}_{3\times3} \Delta t & \mathbf{0} \\
            \mathbf{0} &   \mathbf{0} &  \mathbf{0} & \mathbf{I}_{3\times3} \Delta t \\
            \mathbf{0} &   \mathbf{0} &  \mathbf{0} & \mathbf{0} 
        \end{bmatrix}.
    \end{aligned}
\end{equation}

\subsection{Backward Propagation for Motion Compensation}
\label{sec:lio_back}
Because LiDAR points within a scan are sampled at different time instants $\rho_j \le t_k$, directly associating them with the scan-end state introduces motion distortion. 
To compensate for this effect, the propagated state at the scan-end time $t_k$ is taken as the reference, and the relative motion from each point timestamp $\rho_j$ back to $t_k$ is recovered by backward propagation ($\check{\mathbf{x}}_{j-1} = \check{\mathbf{x}}_j \boxplus \left( -\Delta t \mathbf{f} \left(\check{\mathbf{x}}_j, \mathbf{u}_j, \mathbf{0} \right) \right)$). 
The backward propagation is performed at the frequency of the LiDAR point, which is usually much higher than the IMU rate. 
For all the points sampled between two IMU measurements, we use the left IMU measurement as the input in the back propagation.
\begin{equation}
\label{e:back_prop}
    \begin{aligned}
        {}^{I_k}_{I_{j-1}} \check{\mathbf{t}} 
        &= {}^{I_k}_{I_j} \check{\mathbf{t}} - {}^{I_k}_{I_j} \check{\mathbf{v}} \Delta t, \ \ \ \ \ \ \ \ \ \ \ \ \ \ \ \ \ \ \ \ \ \ \ \ \ \ \ \ \ \ \ \ \ \ \ \rm{s.f.}\ \ {}^{I_k}_{I_m} \check{\mathbf{t}} = \mathbf{0}, \\
        {}^{I_k}_{I_{j-1}} \check{\mathbf{v}}
        &= {}^{I_k}_{I_j} \check{\mathbf{v}} - {}^{I_k}_{I_j} \check{\mathbf{R}} \left(\mathbf{a}_{m_{i-1}} - \widehat{\mathbf{b}}_{\mathbf{a}_k}\right) \Delta t - {}^{I_k} \widehat{\mathbf{g}} \Delta t, \ \ \ \rm{s.f.} \ \ {}^{I_k}_{I_m} \check{\mathbf{v}} = {}^G_{I_k}\mathbf{R}^T {}^G_{I_k}\widehat{\mathbf{v}}, \ \ \ {}^{I_k} \widehat{\mathbf{g}}={}^G_{I_k}\mathbf{R}^T {}^G \widehat{\mathbf{g}}, \\
        {}^{I_k}_{I_{j-1}} \check{\mathbf{R}} 
        &= {}^{I_k}_{I_j} \check{\mathbf{R}} \ \rm{Exp} \left(\Delta t \left(\widehat{\mathbf{b}}_{\bm{\omega}_k} - \bm{\omega}_{m_{i-1}} \right)\right), \ \ \ \ \ \ \ \ \ \ \ \ \ \  \rm{s.f.} \ \  {}^{I_k}_{I_m} \check{\mathbf{R}} = \mathbf{I}
    \end{aligned}
\end{equation}
where $\rho_{j-1} \in \left[\tau_{i-1}, \tau_i \right), \Delta t = \rho_j - \rho_{j-1}$, and $\rm{s.f.}$ means "starting from".

The backward propagation will produce a relative pose between time $\rho_j$ and the scan-end time $t_k$, $ {}^{I_k}_{I_j} \check{\mathbf{T}} = \left({}^{I_k}_{I_j} \check{\mathbf{R}}, \  {}^{I_k}_{I_j} \check{\mathbf{t}} \right)$. 
As shown in Figure~\ref{fig:state_propagation}, this relative pose enables us to project the local measurement ${}^{L_j} \mathbf{p}_j$ to scan-end measurement ${}^{L_k} \mathbf{p}_j$ as follows
\begin{equation}
\label{e:deskew_motion}
    \begin{aligned}
        {}^{L_k} \mathbf{p}_j = {}^I_L \mathbf{T}^{-1} \ {}^{I_k}_{I_j} \check{\mathbf{T}} \ {}^I_L \mathbf{T} \ {}^{L_j} \mathbf{p}_j
    \end{aligned}
\end{equation}
where ${}^I_L \mathbf{T}$ is the known LiDAR-IMU extrinsic (see Section~\ref{sec:preli_kinematics}).

\subsection{Point-to-plane Residual Construction}
\label{sec:lio_res}
Based on the accurate uncertainty modeling of the points and planes in Section~\ref{sec:voxel_prob}, we could easily implement the point-to-plane scan match. 
Given a LiDAR point ${}^G \mathbf{p}_i$ predicted in the world frame with the pose prior, we first find which root voxel (with the coarse map resolution) it lies in by its Hash key.
Then, all the contained sub-voxels are polled for a possible match with the point. 
Specifically, let a sub-voxel contains a plane with normal $\mathbf{n}_i$ and center point $\mathbf{q}_i$, we calculate the point-to-plane distance
\begin{equation}
\label{e:p2pl_d}
    \begin{aligned}
        d_i = \mathbf{n}_i^T\left({}^G \mathbf{p}_i - \mathbf{q}_i\right).
    \end{aligned}
\end{equation}
Considering all these uncertainties, we obtain
\begin{equation}
\label{e:p2pl_d_uncertainty}
    \begin{aligned}
        d_i &= d_i^{gt} - \delta_{d_i} =\left(\mathbf{n}_i^{gt} \boxminus \delta_{\mathbf{n}_i} \right)^T\left({}^G \mathbf{p}_i^{gt} - \delta_{{}^G \mathbf{p}_i} - \left(\mathbf{q}_i^{gt} - \delta_{\mathbf{q}_i}\right)\right) \\
        &= \left(\mathbf{n}_i^{gt}\right)^T \left({}^G \mathbf{p}_i^{gt} - \mathbf{q}_i^{gt} - \left(\delta_{{}^G \mathbf{p}_i} - \delta_{\mathbf{q}_i}\right)\right) - \delta_{\mathbf{n}_i}^T\left({}^G \mathbf{p}_i^{gt} - \mathbf{q}_i^{gt} - \left(\delta_{{}^G \mathbf{p}_i} - \delta_{\mathbf{q}_i}\right)\right) \\
        &= \left(\mathbf{n}_i^{gt}\right)^T  \left({}^G \mathbf{p}_i^{gt} - \mathbf{q}_i^{gt} \right) -  \left(\mathbf{n}_i^{gt}\right)^T \left(\delta_{{}^G \mathbf{p}_i} - \delta_{\mathbf{q}_i}\right) - \delta_{\mathbf{n}_i}^T \left({}^G \mathbf{p}_i^{gt} - \mathbf{q}_i^{gt} \right) +  \delta_{\mathbf{n}_i}^T \left(\delta_{{}^G \mathbf{p}_i} - \delta_{\mathbf{q}_i}\right)  \\
        &= \left(\mathbf{n}_i^{gt}\right)^T  \left({}^G \mathbf{p}_i^{gt} - \mathbf{q}_i^{gt} \right) - \left(\left(\mathbf{n}_i^{gt}\right)^T \left(\delta_{{}^G \mathbf{p}_i} - \delta_{\mathbf{q}_i}\right) - \delta_{\mathbf{n}_i}^T \left(\delta_{{}^G \mathbf{p}_i} - \delta_{\mathbf{q}_i}\right) \right) - \delta_{\mathbf{n}_i}^T \left({}^G \mathbf{p}_i^{gt} - \mathbf{q}_i^{gt} \right) \\
        &=  \left(\mathbf{n}_i^{gt}\right)^T  \left({}^G \mathbf{p}_i^{gt} - \mathbf{q}_i^{gt} \right) - \mathbf{n}_i^T\left(\delta_{{}^G \mathbf{p}_i} - \delta_{\mathbf{q}_i}\right) - \delta_{\mathbf{n}_i}^T \left({}^G \mathbf{p}_i^{gt} - \mathbf{q}_i^{gt}  - \left(\delta_{{}^G \mathbf{p}_i} - \delta_{\mathbf{q}_i}\right)\right) +  \delta_{\mathbf{n}_i}^T \left(\delta_{{}^G \mathbf{p}_i} - \delta_{\mathbf{q}_i}\right) \\
        &= \left(\mathbf{n}_i^{gt}\right)^T  \left({}^G \mathbf{p}_i^{gt} - \mathbf{q}_i^{gt} \right) - \mathbf{n}_i^T\left(\delta_{{}^G \mathbf{p}_i} - \delta_{\mathbf{q}_i}\right) - \delta_{\mathbf{n}_i}^T \left({}^G \mathbf{p}_i - \mathbf{q}_i\right) +\mathcal{O}\left(\delta_{\mathbf{n}_i}, \delta_{\mathbf{p}_i}\right) + \mathcal{O}\left(\delta_{\mathbf{n}_i}, \delta_{\mathbf{q}_i}\right) \\
        &\approx \left(\mathbf{n}_i^{gt}\right)^T  \left({}^G \mathbf{p}_i^{gt} - \mathbf{q}_i^{gt} \right) - \mathbf{n}_i^T\left(\delta_{{}^G \mathbf{p}_i} - \delta_{\mathbf{q}_i}\right) - \delta_{\mathbf{n}_i}^T \left({}^G \mathbf{p}_i - \mathbf{q}_i\right), \\
        d_i^{gt} &= \left(\mathbf{n}_i^{gt}\right)^T \left({}^G \mathbf{p}_i^{gt}  - \mathbf{q}_i^{gt} \right) = \mathbf{0} \\
        \to \delta_{d_i} &= \left({}^G \mathbf{p}_i- \mathbf{q}_i\right)^T \delta_{\mathbf{n}_i} + \mathbf{n}_i^T \delta_{{}^G \mathbf{p}_i} - \mathbf{n}_i^T \delta_{\mathbf{q}_i} 
        = \mathbf{J}_{\mathbf{n}_i} \delta_{\mathbf{n}_i} + \mathbf{J}_{\mathbf{q}_i} \delta_{\mathbf{q}_i} + \mathbf{J}_{{}^G \mathbf{p}_i} \delta_{{}^G \mathbf{p}_i}
    \end{aligned}
\end{equation}
which gives 
\begin{equation}
\label{e:p2pl_d_conv}
    d_i \sim \mathcal{N}\left(0, \Sigma_{d_i} \right), \ \ \Sigma_{d_i}=\mathbf{J}_{d_i} \bm{\Sigma}_{\mathbf{n}_i,\mathbf{q}_i, {}^G \mathbf{p}_i} \mathbf{J}_{d_i}^T 
\end{equation}
where $\mathbf{J}_{d_i} = \begin{bmatrix}
    \mathbf{J}_{\mathbf{n}_i} & \mathbf{J}_{\mathbf{q}_i} & \mathbf{J}_{{}^G \mathbf{p}_i}
\end{bmatrix}=\begin{bmatrix}
    \left({}^G \mathbf{p}_i- \mathbf{q}_i\right)^T & -\mathbf{n}_i^T & \mathbf{n}_i^T
\end{bmatrix}, \ \ \bm{\Sigma}_{\mathbf{n}_i,\mathbf{q}_i, {}^G \mathbf{p}_i} = \begin{bmatrix}
    \bm{\Sigma}_{\mathbf{n}_i,\mathbf{q}_i} & \mathbf{0} \\
    \mathbf{0} & \bm{\Sigma}_{{}^G \mathbf{p}_i}
\end{bmatrix}$.

Point-to-plane associations are filtered according to the probabilistic distance model in (\ref{e:p2pl_d_conv}). 
Specifically, a match is accepted only if the measured distance lies within $3\sigma$, where $\sigma = \sqrt{\Sigma_{d_i}}$.
If multiple planes satisfy this condition, the most probable one is selected; otherwise, the point is discarded to avoid false matches introduced by voxel quantization.

\subsection{Iterated Error-State Kalman Filter Update}
\label{sec:lio_ikf}
This subsection presents the iterated error-state update using point-to-plane residual under the IESKF framework. 

Following Section~\ref{sec:lio_res}, construct the $i$-th valid point-to-plane match observation equation
\begin{equation}
\label{e:lio_obs_model}
    \begin{aligned}
        z_i 
        &= \mathbf{h}_i \left( \mathbf{x}_k \right) + v_i
    \end{aligned}
\end{equation}
where $z_i$ is the point-to-plane distance residual in (\ref{e:p2pl_d}), $\mathbf{h}\left(\mathbf{x}_k\right)$ is the observation function and $v_i \sim \left(\mathbf{0}, \mathbf{Q}_i\right)$ is the observation noise.
By substituting the state $\mathbf{x}_k$ (i.e., the sensor pose $\mathbf{T}_k$) into (\ref{e:p2pl_d}) and linearize it at $\widetilde{\mathbf{x}}_k^{\kappa}=\mathbf{0}$, we obtain
\begin{equation}
\label{e:lio_h_linear}
    \begin{aligned}
        \mathbf{h}_i \left( \mathbf{x}_k \right) 
        &= \mathbf{h}_i \left( \widehat{\mathbf{x}}_k^{\kappa} \boxplus \widetilde{\mathbf{x}}_k^{\kappa}\right) 
        = \mathbf{n}_i^T\left({}^G \mathbf{p}_i - \mathbf{q}_i\right) \\
        &= \mathbf{n}_i^T\left(\left({}^G_I \mathbf{R}_k \left({}^I_L \mathbf{R} {}^L\mathbf{p}_i + {}^I_L \mathbf{t}\right) + {}^G_I \mathbf{t}_k\right) - \mathbf{q}_i\right) \\
        &\approx \mathbf{h}_i \left( \widehat{\mathbf{x}}_k^{\kappa} \right) + \mathbf{H}_i^{\kappa} \widetilde{\mathbf{x}}_k^{\kappa}
    \end{aligned}
\end{equation}
where $\mathbf{H}_i^{\kappa} = \left.\frac{\partial \mathbf{h}_i \left( \mathbf{x}_k \right)}{\partial \widetilde{\mathbf{x}}_k^{\kappa}} \right|_{\widetilde{\mathbf{x}}_k^{\kappa}=\mathbf{0}}$. The exact formulation is shown as below
\begin{equation}
\label{e:lio_hx}
    \begin{aligned}
        \mathbf{H}_i^{\kappa} 
        &= \left.\frac{\partial \mathbf{h}_i \left( \mathbf{x}_k \right)}{\partial \widetilde{\mathbf{x}}_k^{\kappa}} \right|_{\widetilde{\mathbf{x}}_k^{\kappa}=\mathbf{0}} = \left.\frac{\partial \mathbf{h}_i \left( \mathbf{x}_k \right)}{\partial {}^G \mathbf{p}_i} \frac{\partial {}^G \mathbf{p}_i}{\partial  \widetilde{\mathbf{x}}_k^{\kappa}}\right|_{\widetilde{\mathbf{x}}_k^{\kappa}=\mathbf{0}} \\
        &= \mathbf{n}_i^T \left.\frac{\partial {}^G_I \mathbf{R}_k \left({}^I_L \mathbf{R} {}^L\mathbf{p}_i + {}^I_L \mathbf{t}\right) + {}^G_I \mathbf{t}_k}{\partial \widetilde{\mathbf{x}}_k^{\kappa}}\right|_{\widetilde{\mathbf{x}}_k^{\kappa}=\mathbf{0}} \\
        &= \mathbf{n}_i^T \begin{bmatrix}
           - {}^G_I \widehat{\mathbf{R}}_k^{\kappa} \lfloor{}^I_L \mathbf{R} {}^L\mathbf{p}_i + {}^I_L \mathbf{t}\rfloor_{\wedge} & \mathbf{I}_{3\times3} & \mathbf{0}_{3\times3} & \mathbf{0}_{3\times3} & \mathbf{0}_{3\times3} & \mathbf{0}_{3\times3} 
        \end{bmatrix} \\
        &= \begin{bmatrix}
           - \mathbf{n}_i^T{}^G_I \widehat{\mathbf{R}}_k^{\kappa} \lfloor{}^I_L \mathbf{R} {}^L\mathbf{p}_i + {}^I_L \mathbf{t}\rfloor_{\wedge} & \mathbf{n}_i^T & \mathbf{0}_{1\times3} & \mathbf{0}_{1\times3} & \mathbf{0}_{1\times3} & \mathbf{0}_{1\times3} 
        \end{bmatrix}\\
        &= \begin{bmatrix}
           \left( \lfloor{}^I_L \mathbf{R} {}^L\mathbf{p}_i + {}^I_L \mathbf{t}\rfloor_{\wedge} \left({}^G_I \widehat{\mathbf{R}}_k^{\kappa} \right)^T \mathbf{n}_i \right)^T & \mathbf{n}_i^T & \mathbf{0}_{1\times3} & \mathbf{0}_{1\times3} & \mathbf{0}_{1\times3} & \mathbf{0}_{1\times3} 
        \end{bmatrix}.
    \end{aligned}
\end{equation}
Similarly, we can get the observation noise $v_i$ (comes from the uncertainty of point and plane) following the derivation in (\ref{e:p2pl_d_uncertainty})
\begin{equation}
\label{e:lio_obs_noise}
    \begin{aligned}
        v_i 
        &= \mathbf{J}_{\mathbf{n}_i} \delta_{\mathbf{n}_i} + \mathbf{J}_{\mathbf{q}_i} \delta_{\mathbf{q}_i} + \mathbf{J}_{{}^L \mathbf{p}_i} \delta_{{}^L \mathbf{p}_i}, \ \ \
        \mathbf{Q}_i = \mathbf{J}_{v_i} \bm{\Sigma}_{\mathbf{n}_i,\mathbf{q}_i, {}^L \mathbf{p}_i} \mathbf{J}_{v_i}^T, \ \ \
        \bm{\Sigma}_{\mathbf{n}_i,\mathbf{q}_i, {}^L \mathbf{p}_i} 
        = \begin{bmatrix} \bm{\Sigma}_{\mathbf{n}_i,\mathbf{q}_i} & \mathbf{0} \\ \mathbf{0} & \bm{\Sigma}_{{}^L \mathbf{p}_i} \end{bmatrix}, \\
        \mathbf{J}_{v_i} 
        &= \begin{bmatrix} \left(\left({}^G_I \widehat{\mathbf{R}}_k^{\kappa} \left({}^I_L \mathbf{R} {}^L\mathbf{p}_i + {}^I_L \mathbf{t}\right) + {}^G_I \widehat{\mathbf{t}}_k^{\kappa}\right)- \mathbf{q}_i\right)^T & -\mathbf{n}_i^T & \mathbf{n}_i^T {}^G_I \widehat{\mathbf{R}}_k^{\kappa} {}^I_L \mathbf{R} \end{bmatrix}.
    \end{aligned}
\end{equation}
Following (\ref{e:lio_obs_model}) and (\ref{e:lio_hx}), the point-to-plane residual observation model can be obtained
\begin{equation}
\label{e:lio_obs_exact}
    0 \approx  \mathbf{h}_i \left( \widehat{\mathbf{x}}_k^{\kappa} \right) + \mathbf{H}_i^{\kappa} \widetilde{\mathbf{x}}_k^{\kappa} + v_i
\end{equation}

Notice that the prior distribution of $\mathbf{x}_k$ obtained from the forward propagation in Section~\ref{sec:lio_forward} is for 
\begin{equation}
\label{e:lio_prior_jac}
    \begin{aligned}
        \mathbf{x}_k \boxminus \widehat{\mathbf{x}}_k = \left( \widehat{\mathbf{x}}_k^{\kappa} \boxplus \widetilde{\mathbf{x}}_k^{\kappa} \right) \boxminus  \widehat{\mathbf{x}}_k \approx \widehat{\mathbf{x}}_k^{\kappa} \boxminus \widehat{\mathbf{x}}_k + \mathbf{J}^{\kappa} \widetilde{\mathbf{x}}_k^{\kappa}, \\
        \widetilde{\mathbf{x}}_k^{\kappa} \sim \left(\mathbf{0}, \left(\mathbf{J}^{\kappa}\right)^{-1}\widehat{\mathbf{P}}_k\left(\mathbf{J}^{\kappa}\right)^{-T}\right) 
    \end{aligned}
\end{equation}
where $\mathbf{J}^{\kappa} = \left. \frac{\partial \left( \widehat{\mathbf{x}}_k^{\kappa} \boxplus \widetilde{\mathbf{x}}_k^{\kappa} \right) \boxminus  \widehat{\mathbf{x}}_k}{\partial \widetilde{\mathbf{x}}_k^{\kappa}} \right|_{\widetilde{\mathbf{x}}_k^{\kappa}=\mathbf{0}}$. 
For errors other than $\delta \bm{\theta}$ in rotation, $\mathbf{I}$ can be easily obtained.
Next, derive the partial differentiation about $\delta \bm{\theta}$ 
\begin{equation}
\label{e:partial_rot}
    \begin{aligned}
        \frac{\partial \left( \widehat{\mathbf{R}}_k^{\kappa} \boxplus \delta \bm{\theta}_k^{\kappa} \right) \boxminus \widehat{\mathbf{R}}_k}{\partial \delta \bm{\theta}_k^{\kappa}}
        &= \frac{\partial \rm{Log} \left( \widehat{\mathbf{R}}_k^T \widehat{\mathbf{R}}_k^{\kappa}\rm{Exp}\left( \delta \bm{\theta}_k^{\kappa}\right) \right)}{\partial \delta \bm{\theta}_k^{\kappa}} \\
        &\overset{BCH}{=} \frac{\partial \mathcal{J}_l \left(\widehat{\mathbf{R}}_k^{\kappa} \boxminus \widehat{\mathbf{R}}_k \right)^{-T} \delta \bm{\theta}_k^{\kappa} + \widehat{\mathbf{R}}_k^{\kappa} \boxminus \widehat{\mathbf{R}}_k }{\partial \delta \bm{\theta}_k^{\kappa}} \\
        &= \mathcal{J}_l \left(\widehat{\mathbf{R}}_k^{\kappa} \boxminus \widehat{\mathbf{R}}_k \right)^{-T}.
    \end{aligned}
\end{equation}
Thus, the full formulation is
\begin{equation}
\label{e:lio_prior_jac_full}
    \mathbf{J}^{\kappa} = \begin{bmatrix}
        \mathcal{J}_l \left(\widehat{\mathbf{R}}_k^{\kappa} \boxminus \widehat{\mathbf{R}}_k \right)^{-T} & \mathbf{0}_{3\times15} \\
        \mathbf{0}_{15\times3} & \mathbf{I}_{15\times15}
    \end{bmatrix}
\end{equation}
where $\mathcal{J}_l$ defined in (\ref{e:manifold_rules_so3_b}).

Combined the prior in (\ref{e:lio_prior_jac}) with the posteriori distribution from (\ref{e:lio_obs_exact}) yields the maximum a-posteriori estimate (MAP)
\begin{equation}
\label{e:lio_map}
    \min_{\widetilde{\mathbf{x}}_{k}^\kappa} \left( \| \mathbf{x}_k \boxminus \widehat{\mathbf{x}}_k \|^2_{\widehat{\mathbf{P}}^{-1}_k} + \sum_{i=1}^{N} \| \mathbf{h}_i \left( \widehat{\mathbf{x}}_k^{\kappa} \right) + \mathbf{H}_i^{\kappa} \widetilde{\mathbf{x}}_k^{\kappa} \|^2_{\mathbf{Q}^{-1}_i} \right)
\end{equation}
where $\| \mathbf{x} \|_{\mathbf{M}}^2 = \mathbf{x}^T \mathbf{M} \mathbf{x}$. Let $\mathbf{x}_k \boxminus \widehat{\mathbf{x}}_k = \widehat{\mathbf{x}}_k^{\kappa} \boxminus \widehat{\mathbf{x}}_k + \mathbf{J}^{\kappa} \widetilde{\mathbf{x}}_k^{\kappa} = \mathbf{X} + \mathbf{J}^{\kappa} \widetilde{\mathbf{x}}_k^{\kappa}$ and $d_i^{\kappa} = \mathbf{h}_i \left( \widehat{\mathbf{x}}_k^{\kappa} \right)$, the MAP can be represented as 
\begin{equation}
\label{e:lio_map_abb}
    \begin{aligned}
        \mathbf{d}^{\kappa} &= \begin{bmatrix}
            d_1^{\kappa} \\ d_2^{\kappa} \\ \vdots \\ d_N^{\kappa}
        \end{bmatrix}, \ \ \mathbf{H}^{\kappa} = \begin{bmatrix}
            \mathbf{H}_1^{\kappa} \\ \mathbf{H}_2^{\kappa} \\ \vdots \\ \mathbf{H}_N^{\kappa}
        \end{bmatrix}, \ \ 
        \mathbf{Q}_k = \begin{bmatrix}
            \mathbf{Q}_1 & \mathbf{0} & \cdots & \mathbf{0} \\
            \mathbf{0} & \mathbf{Q}_2 & \cdots & \mathbf{0} \\
            \vdots & \vdots &   \ddots &\mathbf{0} \\
             \mathbf{0} & \mathbf{0} &   \cdots & \mathbf{Q}_N
        \end{bmatrix}, \\
        \mathcal{E} 
        &= \min_{\widetilde{\mathbf{x}}_{k}^\kappa} \frac{1}{2} 
            \begin{bmatrix}\mathbf{d}^{\kappa} +  \mathbf{H}^{\kappa} \widetilde{\mathbf{x}}_k^{\kappa} \\ \mathbf{X} + \mathbf{J}^{\kappa} \widetilde{\mathbf{x}}_k^{\kappa} \end{bmatrix}^T 
            \begin{bmatrix} \mathbf{Q}_k & \mathbf{0} \\ \mathbf{0} & \widehat{\mathbf{P}}_k \end{bmatrix}^{-1} 
            \begin{bmatrix}\mathbf{d}^{\kappa} +  \mathbf{H}^{\kappa} \widetilde{\mathbf{x}}_k^{\kappa} \\ \mathbf{X} + \mathbf{J}^{\kappa} \widetilde{\mathbf{x}}_k^{\kappa} \end{bmatrix} = \min_{\widetilde{\mathbf{x}}_{k}^\kappa} \frac{1}{2} \left(\mathbf{M}^{\kappa}\right)^T \mathbf{M}^{\kappa}, \\
        \mathbf{M}^{\kappa} &= \mathbf{S}\begin{bmatrix}\mathbf{d}^{\kappa} +  \mathbf{H}^{\kappa} \widetilde{\mathbf{x}}_k^{\kappa} \\ \mathbf{X} + \mathbf{J}^{\kappa} \widetilde{\mathbf{x}}_k^{\kappa} \end{bmatrix}, \ \ \ \mathbf{S}^T \mathbf{S} = \begin{bmatrix} \mathbf{Q}_k & \mathbf{0} \\ \mathbf{0} & \widehat{\mathbf{P}}_k \end{bmatrix}^{-1}.
    \end{aligned}
\end{equation}
Using Gauss-Newton to solve the MAP, 
\begin{equation}
\label{e:lio_map_solver}
    \begin{aligned}
        \mathbf{J}_{\mathbf{M}}^{\kappa} &= \frac{\partial \mathbf{M}^{\kappa}}{\partial \widetilde{\mathbf{x}}_k^{\kappa}} =  \mathbf{S} \begin{bmatrix}
            \mathbf{H}^{\kappa} \\  \mathbf{J}^{\kappa} \end{bmatrix} , \\
        \delta \widetilde{\mathbf{x}}_k^{\kappa} &= -\left(\left(\mathbf{J}_{\mathbf{M}}^{\kappa}\right)^T\mathbf{J}_{\mathbf{M}}^{\kappa}\right)^{-1} \left(\mathbf{J}_{\mathbf{M}}^{\kappa}\right)^T \mathbf{M}^{\kappa}
    \end{aligned}
\end{equation}
We can use the $\rm{SMW}$ rule $\left(\mathbf{A}^{-1}+\mathbf{B}\mathbf{D}^{-1}\mathbf{C}\right)^{-1}=\mathbf{A}-\mathbf{AB}\left(\mathbf{D+CAB}\right)^{-1}\mathbf{CA}$ to formulate the equation $\left(\left(\mathbf{J}_{\mathbf{M}}^{\kappa}\right)^T\mathbf{J}_{\mathbf{M}}^{\kappa}\right)^{-1}$

\begin{equation}
\label{e:JJ_inv}
    \begin{aligned}
        & \mathbf{A}^{-1} = \left(\mathbf{J}^{\kappa}\right)^T \widehat{\mathbf{P}}_k^{-1} \mathbf{J}^{\kappa}, 
        \mathbf{B} = \left(\mathbf{H}^{\kappa}\right)^T, 
        \mathbf{D}^{-1} =  \mathbf{Q}_k^{-1}, 
        \mathbf{C} = \mathbf{H}^{\kappa}, \\
        & \to \left(\left(\mathbf{J}_{\mathbf{M}}^{\kappa}\right)^T\mathbf{J}_{\mathbf{M}}^{\kappa}\right)^{-1} 
        = \left(\left(\mathbf{J}^{\kappa}\right)^T \widehat{\mathbf{P}}_k^{-1} \mathbf{J}^{\kappa} +  \left(\mathbf{H}^{\kappa}\right)^T\mathbf{Q}_k^{-1} \mathbf{H}^{\kappa} \right)^{-1} \\
        &= \left( \left(\mathbf{J}^{\kappa}\right)^T \widehat{\mathbf{P}}_k^{-1} \mathbf{J}^{\kappa} \right)^{-1}
        - \left( \left(\mathbf{J}^{\kappa}\right)^T \widehat{\mathbf{P}}_k^{-1} \mathbf{J}^{\kappa} \right)^{-1} \left(\mathbf{H}^{\kappa}\right)^T \left(\mathbf{Q}_k + \mathbf{H}^{\kappa} \left( \left(\mathbf{J}^{\kappa}\right)^T \widehat{\mathbf{P}}_k^{-1} \mathbf{J}^{\kappa} \right)^{-1} \left(\mathbf{H}^{\kappa}\right)^T  \right)^{-1} \mathbf{H}^{\kappa} \left( \left(\mathbf{J}^{\kappa}\right)^T \widehat{\mathbf{P}}_k^{-1} \mathbf{J}^{\kappa} \right)^{-1} \\
        &= \left( \mathbf{I} - \left( \left(\mathbf{J}^{\kappa}\right)^T \widehat{\mathbf{P}}_k^{-1} \mathbf{J}^{\kappa} \right)^{-1} \left(\mathbf{H}^{\kappa}\right)^T \left(\mathbf{Q}_k + \mathbf{H}^{\kappa} \left( \left(\mathbf{J}^{\kappa}\right)^T \widehat{\mathbf{P}}_k^{-1} \mathbf{J}^{\kappa} \right)^{-1} \left(\mathbf{H}^{\kappa}\right)^T  \right)^{-1} \mathbf{H}^{\kappa} \right)  \left( \left(\mathbf{J}^{\kappa}\right)^T \widehat{\mathbf{P}}_k^{-1} \mathbf{J}^{\kappa} \right)^{-1} \\
        &= \left( \mathbf{I} - \mathbf{U} \left(\mathbf{H}^{\kappa}\right)^T \left(\mathbf{Q}_k + \mathbf{H}^{\kappa} \mathbf{U}\left(\mathbf{H}^{\kappa}\right)^T  \right)^{-1} \mathbf{H}^{\kappa} \right) \mathbf{U} \\
        &= \left( \mathbf{I} - \mathbf{K} \mathbf{H}^{\kappa} \right) \mathbf{U}
    \end{aligned}
\end{equation}
where $\mathbf{U}= \left( \left(\mathbf{J}^{\kappa}\right)^T \widehat{\mathbf{P}}_k^{-1} \mathbf{J}^{\kappa} \right)^{-1}$ and $\mathbf{K}=\mathbf{U} \left(\mathbf{H}^{\kappa}\right)^T \left(\mathbf{Q}_k + \mathbf{H}^{\kappa} \mathbf{U}\left(\mathbf{H}^{\kappa}\right)^T  \right)^{-1}$.
Combined $\mathbf{K}$ with (\ref{e:JJ_inv}), we can obtain
\begin{equation}
\label{e:JJ_inv1}
    \begin{aligned}
        \mathbf{K} 
        &=\mathbf{U} \left(\mathbf{H}^{\kappa}\right)^T \left(\mathbf{Q}_k + \mathbf{H}^{\kappa} \mathbf{U}\left(\mathbf{H}^{\kappa}\right)^T  \right)^{-1}, \\
        \to \mathbf{U} \left(\mathbf{H}^{\kappa}\right)^T  
        &= \mathbf{K}\left(\mathbf{Q}_k + \mathbf{H}^{\kappa} \mathbf{U}\left(\mathbf{H}^{\kappa}\right)^T  \right) \\
        &= \mathbf{K}\mathbf{Q}_k +  \mathbf{K} \mathbf{H}^{\kappa} \mathbf{U}\left(\mathbf{H}^{\kappa}\right)^T, \\
        \to \mathbf{0} 
        &= \mathbf{K}\mathbf{Q}_k +  \mathbf{K} \mathbf{H}^{\kappa} \mathbf{U}\left(\mathbf{H}^{\kappa}\right)^T - \mathbf{U} \left(\mathbf{H}^{\kappa}\right)^T \\
        &= \mathbf{K}\mathbf{Q}_k - \left(\mathbf{I} -  \mathbf{K} \mathbf{H}^{\kappa}\right)\mathbf{U} \left(\mathbf{H}^{\kappa}\right)^T, \\
        \to \mathbf{K}\mathbf{Q}_k &= \left(\mathbf{I} -  \mathbf{K} \mathbf{H}^{\kappa}\right)\mathbf{U} \left(\mathbf{H}^{\kappa}\right)^T, \\
        \to \left(\left(\mathbf{J}_{\mathbf{M}}^{\kappa}\right)^T\mathbf{J}_{\mathbf{M}}^{\kappa}\right)^{-1} 
        &= \left( \mathbf{I} - \mathbf{K} \mathbf{H}^{\kappa} \right) \mathbf{U} = \mathbf{K}\mathbf{Q}_k \left(\mathbf{H}^{\kappa}\right)^{-T}.
    \end{aligned}
\end{equation}
The rest equation $\left(\mathbf{J}_{\mathbf{M}}^{\kappa}\right)^T \mathbf{M}^{\kappa}$ is derived as
\begin{equation}
\label{e:JM}
    \begin{aligned}
        & \left(\mathbf{J}_{\mathbf{M}}^{\kappa}\right)^T \mathbf{M}^{\kappa} 
        = \begin{bmatrix}
            \left(\mathbf{H}^{\kappa}\right)^{T} &  \left(\mathbf{J}^{\kappa}\right)^T \end{bmatrix} \mathbf{S}^T \mathbf{S}\begin{bmatrix}\mathbf{d}^{\kappa} +  \mathbf{H}^{\kappa} \widetilde{\mathbf{x}}_k^{\kappa} \\ \mathbf{X} + \mathbf{J}^{\kappa} \widetilde{\mathbf{x}}_k^{\kappa} \end{bmatrix} \\
        &= \begin{bmatrix}
            \left(\mathbf{H}^{\kappa}\right)^{T} &  \left(\mathbf{J}^{\kappa}\right)^T \end{bmatrix} 
            \begin{bmatrix} \mathbf{Q}_k & \mathbf{0} \\ \mathbf{0} & \widehat{\mathbf{P}}_k \end{bmatrix}^{-1}
            \begin{bmatrix}\mathbf{d}^{\kappa} +  \mathbf{H}^{\kappa} \widetilde{\mathbf{x}}_k^{\kappa} \\ \mathbf{X} + \mathbf{J}^{\kappa} \widetilde{\mathbf{x}}_k^{\kappa} \end{bmatrix} \\
        &= \left(\mathbf{H}^{\kappa}\right)^{T}\mathbf{Q}_k^{-1} \left(\mathbf{d}^{\kappa} +  \mathbf{H}^{\kappa} \widetilde{\mathbf{x}}_k^{\kappa} \right) + \left(\mathbf{J}^{\kappa}\right)^T \widehat{\mathbf{P}}_k^{-1} \left(\mathbf{X} + \mathbf{J}^{\kappa} \widetilde{\mathbf{x}}_k^{\kappa} \right).
    \end{aligned}
\end{equation}
Following (\ref{e:lio_map_solver}), (\ref{e:JJ_inv1}) and (\ref{e:JM}), yields
\begin{equation}
\label{e:lio_update_x}
    \begin{aligned}
        \delta \widetilde{\mathbf{x}}_k^{\kappa} &= -\left(\left(\mathbf{J}_{\mathbf{M}}^{\kappa}\right)^T\mathbf{J}_{\mathbf{M}}^{\kappa}\right)^{-1} \left(\mathbf{J}_{\mathbf{M}}^{\kappa}\right)^T \mathbf{M}^{\kappa} \\
        &= - \mathbf{K}\mathbf{Q}_k \left(\mathbf{H}^{\kappa}\right)^{-T} \left( \left(\mathbf{H}^{\kappa}\right)^{T}\mathbf{Q}_k^{-1} \left(\mathbf{d}^{\kappa} +  \mathbf{H}^{\kappa} \widetilde{\mathbf{x}}_k^{\kappa} \right) + \left(\mathbf{J}^{\kappa}\right)^T \widehat{\mathbf{P}}_k^{-1} \left(\mathbf{X} + \mathbf{J}^{\kappa} \widetilde{\mathbf{x}}_k^{\kappa} \right) \right) \\
        &= -\mathbf{K}  \left(\mathbf{d}^{\kappa} +  \mathbf{H}^{\kappa} \widetilde{\mathbf{x}}_k^{\kappa} \right) - \mathbf{K}\mathbf{Q}_k\left(\mathbf{H}^{\kappa}\right)^{-T}\left(\mathbf{J}^{\kappa}\right)^T \widehat{\mathbf{P}}_k^{-1} \left(\mathbf{X} + \mathbf{J}^{\kappa} \widetilde{\mathbf{x}}_k^{\kappa} \right) \\
        &= -\mathbf{K}  \left(\mathbf{d}^{\kappa} +  \mathbf{H}^{\kappa} \widetilde{\mathbf{x}}_k^{\kappa} \right) - \left(\mathbf{I} -  \mathbf{K} \mathbf{H}^{\kappa}\right)\mathbf{U} \left(\mathbf{H}^{\kappa}\right)^T\left(\mathbf{H}^{\kappa}\right)^{-T}\left(\mathbf{J}^{\kappa}\right)^T \widehat{\mathbf{P}}_k^{-1} \left(\mathbf{X} + \mathbf{J}^{\kappa} \widetilde{\mathbf{x}}_k^{\kappa} \right) \\
        &= -\mathbf{K}  \left(\mathbf{d}^{\kappa} +  \mathbf{H}^{\kappa} \widetilde{\mathbf{x}}_k^{\kappa} \right) - \left(\mathbf{I} -  \mathbf{K} \mathbf{H}^{\kappa}\right)
        \left( \left(\mathbf{J}^{\kappa}\right)^T \widehat{\mathbf{P}}_k^{-1} \mathbf{J}^{\kappa} \right)^{-1}
        \left(\mathbf{J}^{\kappa}\right)^T \widehat{\mathbf{P}}_k^{-1} \left(\mathbf{X} + \mathbf{J}^{\kappa} \widetilde{\mathbf{x}}_k^{\kappa} \right) \\
        &= -\mathbf{K}  \left(\mathbf{d}^{\kappa} +  \mathbf{H}^{\kappa} \widetilde{\mathbf{x}}_k^{\kappa} \right) - \left(\mathbf{I} -  \mathbf{K} \mathbf{H}^{\kappa}\right)
        \left(\mathbf{J}^{\kappa}\right)^{-1} \widehat{\mathbf{P}}_k \left(\mathbf{J}^{\kappa} \right)^{-T}
        \left(\mathbf{J}^{\kappa}\right)^T \widehat{\mathbf{P}}_k^{-1} \left(\mathbf{X} + \mathbf{J}^{\kappa} \widetilde{\mathbf{x}}_k^{\kappa} \right) \\
        &= -\mathbf{K}  \left(\mathbf{d}^{\kappa} +  \mathbf{H}^{\kappa} \widetilde{\mathbf{x}}_k^{\kappa} \right) - \left(\mathbf{I} -  \mathbf{K} \mathbf{H}^{\kappa}\right)
        \left(\mathbf{J}^{\kappa}\right)^{-1}  \left(\mathbf{X} + \mathbf{J}^{\kappa} \widetilde{\mathbf{x}}_k^{\kappa} \right) \\
        &= -\mathbf{K}\mathbf{d}^{\kappa} - \mathbf{K}\mathbf{H}^{\kappa} \widetilde{\mathbf{x}}_k^{\kappa} - \left(\mathbf{I} -  \mathbf{K} \mathbf{H}^{\kappa}\right) \left(\mathbf{J}^{\kappa}\right)^{-1}  \mathbf{X} - \left(\mathbf{I} -  \mathbf{K} \mathbf{H}^{\kappa}\right) \left(\mathbf{J}^{\kappa}\right)^{-1} \mathbf{J}^{\kappa} \widetilde{\mathbf{x}}_k^{\kappa} \\
        &= -\widetilde{\mathbf{x}}_k^{\kappa} -\mathbf{K}\mathbf{d}^{\kappa} - \left(\mathbf{I} -  \mathbf{K} \mathbf{H}^{\kappa}\right) \left(\mathbf{J}^{\kappa}\right)^{-1} \left(\widehat{\mathbf{x}}_k^{\kappa} \boxminus \widehat{\mathbf{x}}_k \right), \\
        \left(\widetilde{\mathbf{x}}_k^{\kappa}\right)^{*}
        &= \widetilde{\mathbf{x}}_k^{\kappa} \boxplus \delta \widetilde{\mathbf{x}}_k^{\kappa} 
        = -\mathbf{K}\mathbf{d}^{\kappa} - \left(\mathbf{I} -  \mathbf{K} \mathbf{H}^{\kappa}\right) \left(\mathbf{J}^{\kappa}\right)^{-1} \left(\widehat{\mathbf{x}}_k^{\kappa} \boxminus \widehat{\mathbf{x}}_k \right), \\
        \widehat{\mathbf{x}}_k^{\kappa+1} &=  \widehat{\mathbf{x}}_k^{\kappa} \boxplus \left( -\mathbf{K}\mathbf{d}^{\kappa} - \left(\mathbf{I} -  \mathbf{K} \mathbf{H}^{\kappa}\right) \left(\mathbf{J}^{\kappa}\right)^{-1} \left(\widehat{\mathbf{x}}_k^{\kappa} \boxminus \widehat{\mathbf{x}}_k \right) \right).
    \end{aligned}
\end{equation}
Thus, the updated estimate $\widehat{\mathbf{x}}_{k}^{\kappa+1}$ is then used to compute the residual in Section~\ref{sec:lio_res} and repeat the process until convergence {(i.e., $\| \widehat{\mathbf{x}}_{k}^{\kappa+1} \boxminus \widehat{\mathbf{x}}_{k}^\kappa\| < \epsilon$)}. 
After convergence, the optimal state estimation and covariance is
\begin{equation}
\label{e:state_update}
    \begin{aligned}
        \bar{\mathbf{x}}_{k} &= \widehat{\mathbf{x}}_{k}^{\kappa+1},\ \bar{\mathbf{P}}_k = \left( \mathbf{I} - \mathbf{K} \mathbf{H}^{\kappa} \right) \mathbf{U}
    \end{aligned}
\end{equation}

In~\cite{fastlio2}, a new Kalman gain formula which is different from $\mathbf{K} =\mathbf{U} \left(\mathbf{H}^{\kappa}\right)^T \left(\mathbf{Q}_k + \mathbf{H}^{\kappa} \mathbf{U}\left(\mathbf{H}^{\kappa}\right)^T  \right)^{-1}$ greatly saves the computation as the state dimension is usually much lower than measurements in LIO (e.g., more than $1000$ effective feature points in a scan for $10$ Hz scan rate while the state dimension is only $18$).
Using the $\rm{SMW}$ rule $\left(\mathbf{A}^{-1}+\mathbf{B}\mathbf{D}^{-1}\mathbf{C}\right)^{-1}=\mathbf{A}-\mathbf{AB}\left(\mathbf{D+CAB}\right)^{-1}\mathbf{CA}$ to formulate the new Kalman gain formula
\begin{equation}
\label{e:kalman_gain_new}
    \begin{aligned}
        \mathbf{A}^{-1} &= \mathbf{Q}_k, \mathbf{B} = \mathbf{H}^{\kappa},  \mathbf{D}^{-1} = \mathbf{U}, \mathbf{C} = \left(\mathbf{H}^{\kappa}\right)^T\\
        \to \mathbf{K} 
        &=\mathbf{U} \left(\mathbf{H}^{\kappa}\right)^T \left(\mathbf{Q}_k + \mathbf{H}^{\kappa} \mathbf{U}\left(\mathbf{H}^{\kappa}\right)^T  \right)^{-1} \\
        &=\mathbf{U} \left(\mathbf{H}^{\kappa}\right)^T \left(\mathbf{Q}_k^{-1} - \mathbf{Q}_k^{-1} \mathbf{H}^{\kappa} \left(\mathbf{U}^{-1}+\left(\mathbf{H}^{\kappa}\right)^T \mathbf{Q}_k^{-1}\mathbf{H}^{\kappa} \right)^{-1} \left(\mathbf{H}^{\kappa}\right)^T\mathbf{Q}_k^{-1} \right) \\
        &=\mathbf{U} \left(\mathbf{H}^{\kappa}\right)^T \left(\mathbf{I} - \mathbf{Q}_k^{-1} \mathbf{H}^{\kappa} \left(\mathbf{U}^{-1}+\left(\mathbf{H}^{\kappa}\right)^T \mathbf{Q}_k^{-1}\mathbf{H}^{\kappa} \right)^{-1} \left(\mathbf{H}^{\kappa}\right)^T \right) \mathbf{Q}_k^{-1} \\
        &= \left(\mathbf{U} \left(\mathbf{H}^{\kappa}\right)^T - \mathbf{U} \left(\mathbf{H}^{\kappa}\right)^T\mathbf{Q}_k^{-1} \mathbf{H}^{\kappa} \left(\mathbf{U}^{-1}+\left(\mathbf{H}^{\kappa}\right)^T \mathbf{Q}_k^{-1}\mathbf{H}^{\kappa} \right)^{-1} \left(\mathbf{H}^{\kappa}\right)^T \right) \mathbf{Q}_k^{-1} \\
        &= \left(\mathbf{U} \left(\mathbf{H}^{\kappa}\right)^T - 
        \left(\mathbf{U} \left( \mathbf{U}^{-1} + \left(\mathbf{H}^{\kappa}\right)^T\mathbf{Q}_k^{-1} \mathbf{H}^{\kappa} \right) -\mathbf{I} \right)
        \left(\mathbf{U}^{-1}+\left(\mathbf{H}^{\kappa}\right)^T \mathbf{Q}_k^{-1}\mathbf{H}^{\kappa} \right)^{-1} \left(\mathbf{H}^{\kappa}\right)^T \right) \mathbf{Q}_k^{-1} \\
        &= \left(\mathbf{U} \left(\mathbf{H}^{\kappa}\right)^T - 
        \mathbf{U} \left(\mathbf{H}^{\kappa}\right)^T + \left(\mathbf{U}^{-1}+\left(\mathbf{H}^{\kappa}\right)^T \mathbf{Q}_k^{-1}\mathbf{H}^{\kappa} \right)^{-1} \left(\mathbf{H}^{\kappa}\right)^T \right) \mathbf{Q}_k^{-1} \\
        &= \left(\mathbf{U}^{-1}+\left(\mathbf{H}^{\kappa}\right)^T \mathbf{Q}_k^{-1}\mathbf{H}^{\kappa} \right)^{-1} \left(\mathbf{H}^{\kappa}\right)^T\mathbf{Q}_k^{-1}.
    \end{aligned}
\end{equation}

In summary, the full state estimation is summarized in Algorithm~\ref{alg:iekf}.
\begin{algorithm}[htbp]
\caption{\textbf{Iterated State Estimation}}
\label{alg:iekf}
    \SetKwInOut{Input}{Input}\SetKwInOut{Output}{Output}\SetKwInOut{Start}{Start}\SetKwInOut{blank}{}
    \Input{Last optimal estimation $\bar{\mathbf{x}}_{k-1}$ and $\bar{\mathbf{P}}_{k-1}$, \\
    IMU inputs ($\mathbf{a}_{m}$, $\bm{\omega}_m$) in current scan; \\
    LiDAR feature points ${}^{L_j}{\mathbf p}_j$ in current scan.}
    Forward propagation to obtain state $\widehat{\mathbf{x}}_k$ via (\ref{e:forward_prop}) and covariance $\widehat{\mathbf{P}}_k$ via (\ref{e:cov_forward});\\
    Backward propagation to obtain ${}^{L_k}{\mathbf{p}}_j$ via (\ref{e:back_prop}) and (\ref{e:deskew_motion});\\
    $\kappa = -1$, $\widehat{\mathbf{x}}_{k}^{\kappa = 0} = \widehat{\mathbf{x}}_{k}$; \\
        \Repeat{$\| \widehat{\mathbf{x}}_{k}^{\kappa+1} \boxminus \widehat{\mathbf{x}}_{k}^\kappa\|< \bm{\epsilon}$}{
        $\kappa =\kappa+1$;\\
        Compute $\mathbf{J}^{\kappa}$ via (\ref{e:lio_prior_jac_full}) and $\mathbf{U} = \left(\mathbf{J}^{\kappa}\right)^{-1} \widehat{\mathbf{P}}_k \left(\mathbf{J}^{\kappa} \right)^{-T} $;\\
        Compute residual $d_i^{\kappa}$ (\ref{e:lio_h_linear}) and Jacobin $\mathbf{H}_i^\kappa$ (\ref{e:lio_hx});\\
        Compute the state update $\widehat{\mathbf{x}}_{k}^{\kappa+1}$ via (\ref{e:lio_update_x}) with the Kalman gain $\mathbf{K}$ from (\ref{e:kalman_gain_new});\\
        }
    $\bar{\mathbf{x}}_k = \widehat{\mathbf{x}}_{k}^{\kappa+1}$; $\bar{\mathbf{P}}_k = \left( \mathbf{I} - \mathbf{K} \mathbf{H}^{\kappa} \right) \mathbf{U}$.\\
    \Output{Current optimal estimation $\bar{\mathbf{x}}_k$ and $\bar{\mathbf{P}}_k$.}
\end{algorithm}


\section{Conclusion}
\label{sec:conclusion}

This note presented a concise mathematical formulation of tightly-coupled LIO under the IESKF framework using a VoxelMap representation. 
By adopting consistent notation and explicitly deriving the associated probabilistic models, the interaction between voxel-based map representation and state estimation was clarified. 
The resulting formulation unifies geometric modeling and filter-based estimation within a coherent framework, and serves as a technical reference for understanding the foundations of VoxelMap-based tightly-coupled LIO systems.

\newpage
\bibliographystyle{IEEEtran}
\bibliography{ref}

\end{document}